\documentclass[journal]{IEEEtran}
\usepackage{algorithm}
\usepackage{algpseudocode}
\usepackage{graphicx} % Add this in your preamble
\usepackage{amsmath}
\usepackage{enumitem} 
\usepackage{tabularx} 
\usepackage{tikz}
\usetikzlibrary{positioning}
\usepackage{booktabs}
\usepackage[colorlinks=true,
    linkcolor=blue,
    citecolor=blue,
    urlcolor=blue]{hyperref}
\usepackage{enumitem}
\begin{document}
\title{Federated Learning for Global Carbon Emission
Forecasting: A Hybrid Time-Series Approach with
Statistical and Neural Models}

%\author{Attia Qammar\textsuperscript{1}\thanks{School of Computing and Artificial Intelligence, Southwest Jiaotong University, Chengdu, 610031, China, e-mail: (q.attia@yahoo.com, a.attia@swjtu.edu.cn)}, {Qazi Haseeb Yousaf\textsuperscript {2}}\thanks{Department of Computer Science, Bahria University, Islamabad. e-mail: (qhyousaf.buic@bahria.edu.pk)}, {Ali Azam}\textsuperscript{3}}\thanks{School of Mechanical Engineering, Southwest Jiaotong University, Chengdu, 610031, PR China, e-mail:(aliazam@swjtu.edu.cn)}, {Ammar Ahmed}\textsuperscript {3}}\thanks{School of Mechanical Engineering, Southwest Jiaotong University, Chengdu, 610031, PR China. e-mail: (ammartahir@swjtu.edu.cn)}}
\author{
Attia Qammar\textsuperscript{1}\thanks{\textsuperscript{1}School of Computing and Artificial Intelligence, Southwest Jiaotong University, Chengdu, 610031, China, e-mail: \texttt{q.attia@yahoo.com}, \texttt{a.attia@swjtu.edu.cn}}, 
Qazi Haseeb Yousaf\textsuperscript{2}\thanks{\textsuperscript{2}Department of Computer Science, Bahria University, Islamabad, e-mail: \texttt{qhyousaf.buic@bahria.edu.pk}}, 
Ali Azam\textsuperscript{3}\thanks{\textsuperscript{3}School of Mechanical Engineering, Southwest Jiaotong University, Chengdu, 610031, PR China, e-mail: \texttt{aliazam@swjtu.edu.cn}}, 
% Ammar Ahmed\textsuperscript{3}\thanks{\textsuperscript{3}School of Mechanical Engineering, Southwest Jiaotong University, Chengdu, 610031, PR China, e-mail: \texttt{ammartahir@swjtu.edu.cn}},
Abdenacer Naouri\textsuperscript{1}\thanks{\textsuperscript{1}School of Computing and Artificial Intelligence, Southwest Jiaotong University, Chengdu, 610031, China, e-mail: \texttt{nacer.naouri@gmail.com}},
Tianrui Li\textsuperscript{1*}\thanks{Corresponding author\textsuperscript{1*}: School of Computing and Artificial Intelligence, Southwest Jiaotong University, Chengdu, 610031, China, e-mail: \texttt{trli@swjtu.edu.cn}}
}

% <-this % stops a space
%

% <-this % stops a space
% \thanks{Manuscript received April 19, 2005; revised August 26, 2015.}}
\maketitle
\begin{abstract}
Climate change, primarily caused by CO\textsubscript{2} emissions, requires precise forecasting instruments for effective mitigation strategies. However, current state-of-the-art models typically rely on centralized data aggregation, which is often impractical due to strict data privacy regulations and the distributed nature of emission data across countries and industrial sectors. This study proposes a novel federated hybrid forecasting architecture for privacy-preserving carbon-emission prediction, integrating ARIMA-based trend modeling, GARCH-based volatility modeling, LSTM-Attention temporal representation learning, and XGBoost prediction within a unified federated learning framework. The novelty of the proposed framework lies in the integration of complementary statistical and neural forecasting components within a privacy-preserving federated environment, enabling the joint modeling of temporal dependencies, emission volatility, and distributed client collaboration without sharing raw data. The framework was evaluated using data from 14 clients and achieved client R² values between 0.50 and 0.97 (average 0.73), RMSE values ranging from 0.06 to 2.35 (average 1.21), and MAPE values between 1.5\% and 11.3\% (average 6.5\%). These results demonstrate that the proposed framework provides a scalable, accurate, and regulation-compliant solution for collaborative carbon-emission forecasting.
\end{abstract}

\begin{IEEEkeywords}
Carbon Emission, time-series forecasting, Federated Learning, LSTM, XGBoost
\end{IEEEkeywords}

\section{Introduction}
\IEEEPARstart{C}{limate} change has become a critical worldwide concern of the 21st century, mostly due to the build-up of greenhouse gases in the atmosphere, with carbon dioxide (CO\textsubscript{2}) being the predominant factor in anthropogenic global warming \cite{ipcc2024, cui2023}. Increasing CO\textsubscript{2} emissions have increased demand of global, political, and scientific initiatives to create reliable forecasting methods for emission trends, as precise predictions are essential for formulating effective mitigation strategies, reconciling economic growth with environmental preservation, and attaining sustainable development objectives \cite{alkheder2022forecasting}.
% New data Added
However, achieving accurate and actionable forecasts is increasingly complicated by the fragmented nature of emission data. Industrial, governmental, and regional institutions gather their own data that is rarely centralized because of privacy, competition, and strict legal frameworks such as the General Data Protection Regulation (GDPR) of the European Union (EU). Such limitations pose an inherent conflict, where the scientific and policy communities have an urgent need of accurate large scale predictions, but the data required to perform that modelling is locked with individual authorities and under legal constraints \cite{9599369}.

China, being the foremost carbon emitter globally, plays a crucial role in the international climate change agenda. The Chinese government has pledged to peak CO\textsubscript{2} emissions by 2030 and attain carbon neutrality by 2060, in accordance with the Paris Agreement \cite{zhao2022china,wang2017}. The ``dual-carbon'' objectives require precise forecasting instruments to inform policy choices, especially in high-emission sectors like industry, energy, and transportation \cite{qiu2023first}. The industrial sector is a primary focus for carbon reduction initiatives due to its significant contribution to national emissions and is highly under legal constraints for data sharing\cite{yang2020}. Therefore, there is an urgent need for such as forecasting system that is not only accurate but have the capability of decentralized processing.
% Forecasting emission trajectories in this industry helps prevent sudden or uncoordinated measures, assuring alignment between environmental objectives, economic stability, and social well-being.

On the other hand, technical advancements with the available data has also evolved in the last decade. Throughout the years, several approaches have been utilised for predicting CO\textsubscript{2} emissions. Econometric methods, including Vector Autoregression (VAR), STIRPAT, and Autoregressive Integrated Moving Average (ARIMA), are extensively employed to analyse time-series correlations \cite{wang2017, yang2020, Appiah2018}. Although these approaches provide statistical validity but they frequently depend on stringent assumptions and possess limited ability to capture nonlinear or highly dynamic patterns in emission data \cite{10423871}. Furthermore, Artificial intelligence (AI) models, such as support vector machines (SVM), deep learning architectures, generalised regression neural networks (GRNN), and optimization-based frameworks like whale optimisation extreme learning machines, have exhibited exceptional efficacy in modelling intricate, nonlinear relationships \cite{dai2018forecasting, Niu2020}. 
% Nonetheless, these models need extensive, centralised datasets and are prone to overfitting and local optima. Also, Grey prediction models, such as fractional grey Riccati models and joint grey algorithms, are proficient in situations characterised by sparse or irregular data, producing robust regular sequences from minimal samples \cite{wang2022grey, gao2021b, gao2021novel}. However, their application is typically confined to short-term or exponential growth forecasts.

% Hybrid models have been devised to enhance prediction accuracy by integrating several methodologies, such as the combination of GM(1,1) with least squares SVM \cite{dai2018} or LSTM networks with grey correlation analysis and principal component analysis \cite{huang2019grey}. These hybrid techniques frequently surpass individual models; yet, they continue to exhibit variable performance across many datasets, and none intrinsically mitigate privacy problems.
% In actuality, carbon emission records from industrial and governmental sources are
A significant constraint of all the aforementioned methods is their reliance on centralised data aggregation. The major reason for that is the already mentioned strict regulations
 often disseminated around many organisations and are subject to stringent privacy, security, and commercial confidentiality regulations \cite{ardakani2023federated}.
 % Legal regulations, such the EU's General Data Protection Regulation (GDPR), put rigorous stipulations on data exchange, rendering centralised training progressively unfeasible
 This difficulty highlights the necessity for approaches that facilitate collaborative model building while safeguarding sensitive data.

Federated Learning (FL) fulfils this requirement by enabling several users to collectively train a shared model while maintaining the locality of raw data. Updates to the model, instead of the original data, are sent to a central server for aggregation. Federated Learning (FL) has exhibited significant potential in various fields, including healthcare \cite{Chen2020, Yu2021}, smart homes \cite{alvodji2019iotfla}, industrial engineering \cite{saputra2019energy}, traffic flow forecasting \cite{liu2020carbon, liu2020privacy}, and energy consumption modelling \cite{zhao2022china, tun2021federated, gholizadeh2022federated}. Despite this broad spectrum of applications, its utilization in carbon emission predictions remains mostly unexplored specifically in terms of hybridization of models.

This study presents a methodology based on federated learning for forecasting CO\textsubscript{2} emissions. The proposed framework is designed to support forecasting of both total and sector-specific carbon emissions. In this study, the framework is evaluated using total carbon-emission data, while its extension to sector-specific forecasting is discussed as a potential application. The study utilizes LSTM for generating the latent feature space and applies machine learning algorithms to finally forecast the expected emission value. Furthermore, in the proposed model for forecasting carbon emissions, ARIMA (AutoRegressive Integrated Moving Average) and GARCH (Generalised AutoRegressive Conditional Heteroskedasticity) perform as robust statistical feature generators. ARIMA identifies the fundamental temporal trends and autocorrelations in historical carbon emission data, enabling the model to incorporate both recent and long-term patterns. Their integration guarantees that LSTM and XGBoost are cognisant of the intrinsic seasonality and trend patterns within the time series. GARCH, conversely, represents the time-dependent volatility of emissions, encapsulating intervals of significant variation and uncertainty. Incorporating ARIMA and GARCH outputs as input features enhances the LSTM and XGBoost models by providing more comprehensive temporal and variability data, hence augmenting their prediction accuracy and resilience, especially amid erratic spikes or volatile emission patterns. This hybrid methodology integrates the interpretability of statistical techniques with the learning capabilities of sophisticated machine learning models.

% Overall, the proposed framework facilitates collaboration among decentralised data sectors without necessitating the centralisation of sensitive information. This study intends to assess the feasibility, prediction accuracy, and potential of this privacy-preserving architecture compared to standard centralised models to inform sustainable environmental policy. This approach tackles a significant research need at the convergence of environmental modelling and privacy-preserving machine learning, providing a scalable and regulation-compliant framework for enhancing carbon emission predictions. 

Despite substantial progress in carbon-emission forecasting and federated learning, existing studies typically focus on statistical forecasting, deep learning forecasting, or federated learning independently. Moreover, existing federated forecasting studies rarely incorporate explicit trend and volatility modeling alongside deep temporal representation learning. To the best of our knowledge, few studies have investigated has been given to the unified integration of ARIMA-based trend extraction, GARCH-based volatility modeling, LSTM-Attention temporal representation learning, and XGBoost forecasting within a privacy-preserving federated carbon-emission forecasting framework.

The major contributions of this study are summarised as follows:

\begin{itemize}

\item This study proposes a federated hybrid statistical-neural forecasting architecture for privacy-preserving carbon-emission prediction, enabling collaborative learning across distributed clients without sharing raw emission data.

\item The proposed framework introduces a unified feature-level integration strategy that combines ARIMA-based trend modeling, GARCH-based volatility modeling, LSTM-Attention temporal representation learning, and XGBoost prediction within a federated learning environment.

\item The proposed architecture integrates complementary statistical and neural forecasting mechanisms to jointly model long-term trends, volatility dynamics, nonlinear temporal dependencies, and cross-feature interactions in carbon-emission time series.

\item A comprehensive evaluation is conducted under multiple data availability settings (30\%, 50\%, and 100\%) and varying federation scales (5, 9, and 14 clients), providing an extensive assessment using R², MSE, MAE, RMSE, and MAPE metrics.

\item The study analyses the effects of federated aggregation on forecasting performance and provides insights into the trade-offs between collaborative learning, privacy preservation, and predictive accuracy.

\end{itemize}

The remainder of this paper is organised as follows. Section~II presents a comprehensive \textit{Literature Study}, reviewing existing econometric, machine learning, grey, hybrid, and privacy-preserving approaches for CO\textsubscript{2} emission forecasting. Section~III outlines the \textit{Methods and Materials}, detailing the dataset, preprocessing steps, ARIMA and GARCH feature generation, and the federated learning architecture employed. Section~IV discusses the \textit{Experimental Setup and Results}, presenting evaluations across different data formations and client scenarios, followed by an in-depth analysis of performance improvements and limitations. Finally, the paper concludes with insights into the applicability of the proposed framework for sustainable environmental modelling and future research directions. 

\section{Literature Study}
The forecasting of carbon dioxide (CO\textsubscript{2}) emissions has been a primary focus of environmental modelling research due to its pivotal role in climate change mitigation. Conventional econometric models, including Vector Autoregression (VAR), STIRPAT, and Autoregressive Integrated Moving Average (ARIMA), have been extensively utilised for modelling emission patterns owing to their statistical robustness \cite{wang2017, yang2020}. Nonetheless, their dependence on linear assumptions and challenges in managing highly nonlinear and dynamic patterns constrain their accuracy in intricate, real-world situations. On the other hand, Artificial intelligence (AI) and Machine Learning (ML) methodologies have demonstrated significant potential for improving predictive accuracy. Techniques including Support Vector Machines (SVM), Deep Learning (DL) architectures, Generalised Regression Neural Networks (GRNN), and optimization-based frameworks such as Extreme Learning Machines have effectively modelled nonlinear relationships in emission data \cite{alkheder2022forecasting, sun2022predictions}. Despite their effectiveness, these approaches sometimes require massive centralized databases, raising concerns about data protection, regulatory compliance, and computing demands.

An increasing amount of research acknowledges that the data necessary for precise carbon emission modelling is frequently dispersed among various entities—industries, governmental agencies, and research institutions—constrained by privacy, security, and commercial confidentiality regulations, including the EU’s GDPR \cite{kairouz2021advances}. This establishes considerable obstacles for centralised learning, highlighting the necessity for decentralised alternatives.
Federated Learning (FL), initially articulated by McMahan et al. \cite{konevcny2016federated}, tackles this issue by facilitating collaborative model training without necessitating the centralisation of raw data. FL has been utilised in healthcare \cite{chen2020fedhealth, hakak2020framework}, smart homes \cite{alvodji2019iotfla}, industrial engineering \cite{saputra2019energy}, traffic flow forecasting \cite{liu2020carbon, liu2024online}, and energy consumption modelling \cite{tun2021federated, zhang2022federated}. 

Recent research has begun to highlight another vital aspect: the carbon impact of machine learning and Deep Learning systems.  However, Deep learning techniques are more resource-intensive, requiring substantial GPU hours for model training in energy-consuming data centers. \cite{qiu2023first}. Between 2012 and 2018, the computational resources employed for cutting-edge machine learning escalated by over 300,000 times \cite{qiu2023first}, with substantial natural language processing models producing as much as 284 tonnes of CO\textsubscript{2}-equivalent emissions \cite{xiong2024}. Data centres together consume around 200 TWh per year, accounting for about 0.3\% of global carbon emissions \cite{ipcc2018global, 9996132}.
Although federated learning provides a privacy-preserving option, its environmental consequences are not well comprehended. The inaugural comprehensive carbon footprint analysis of FL conducted by Qiu et al. \cite{qiu2023first} indicated that emissions resulting from communication overhead between clients and servers may vary from 0.7\% to more than 96\% of total emissions. In certain setups, federated learning (FL) produced much more carbon emissions than centralised training—potentially hundreds of times higher—attributable to the frequent communication cycles and inefficiencies in distributed hardware. This discovery underscores the necessity for carbon-conscious federated learning architectures, wherein hyperparameters, aggregation methodologies (e.g., FedAvg, FedAdam), and deployment configurations are optimised for both performance and energy efficiency. However, it is important to note that the high emissions reported by Qiu et al. \cite{qiu2023first} arise primarily from conventional FL configurations that involve frequent communication rounds, large model updates, and inefficient device synchronization.

Appiah \cite{appiah2018investigating} and Wang \& Lin \cite{wang2017} utilised standard econometric methods, including VAR, ARIMA, and STIRPAT models, for estimating carbon emissions. These approaches offer robust statistical interpretability and theoretical rigour, rendering them advantageous for long-term trend research. Nonetheless, their presumption of linear correlations constrains their capacity to encapsulate intricate nonlinear dynamics, particularly in swiftly evolving contexts.  
AlKheder \& Almusalam \cite{alkheder2022forecasting} advanced the discipline by employing AI-based techniques, including SVM, GRNN, and deep learning architectures. These methods exhibited enhanced precision by identifying complex nonlinear relationships in emission data. Nonetheless, they continue to rely on centralised databases, which presents privacy problems and scalability issues when utilised with remote or sensitive data sources.  

Wang \& Zhang \cite{wang2022grey} and Gao et al. \cite{gao2021prediction, gao2021} advanced grey system modelling by extending grey prediction models to fractional and Riccati-based forms, with good accuracy for tiny and irregular datasets. These methodologies are especially appropriate for short-term and data-deficient situations. Nonetheless, their forecasting effectiveness declines in long-term projections when volatility and exogenous shocks prevail.  
% Dai et al. \cite{dai2018forecasting} amalgamated GM(1,1) with Least Squares Support Vector Machine, whereas
Hybrid models have been developed to mitigate some deficiencies.  Huang et al. \cite{huang2019} fused Long Short-Term Memory with Principal Component Analysis and Grey Correlation Analysis. Hybrid methodologies frequently surpass single-model frameworks for robustness and predictive accuracy. Nevertheless, they inadequately handle privacy issues and are significantly reliant on datasets, resulting in variable performance across different areas.  

Ultimately, the seminal research conducted by McMahan et al. \cite{konevcny2016federated} on Federated Learning (FedAvg) facilitated decentralised machine learning while safeguarding data privacy across clients. Subsequent research, including Qiu et al. \cite{qiu2023first, savazzi2023energy}, particularly analysed the carbon footprint of federated learning, demonstrating that although it offers privacy-preserving advantages, the communication overhead can markedly elevate emissions relative to centralised learning. Table \ref{tab:fl_forecasting} shows the category wise techniques applied in domain of Federated Learning provided in the literature.
% Anew paragraph added to link research gap and our findings.
Critically analyzing the literature, there can be few gaps that needs to be addressed through this research. Firstly, the FL can be adopted as a decentralized solution to the carbon emission problem with an aim to provide privacy and precision at a same time. Further, a research by Cui et al. \cite{cui2023federated} applied SARIMA to capture the variations of time series data using sector-wise data, but their study only emphasizes at single sector. It leads to a critical gap in the research forcing to build separate model for each sector that can be costly as well as consumes a lot of energy for training and communication. Further, linking this with Qiu et al. \cite{qiu2023first} findings that the complexity of communication and model that is associated with federated learning may incur significant carbon overheads, and building on the energy-conscious analyses by Savazzi et al. \cite{savazzi2023energy} and the machine-learning carbon accounting evidence by Xiong et al. \cite{xiong2024}, our framework minimizes the carbon footprint of FL with a number of design decisions. First, the communication size is limited to small LSTM weight updating through the deployment of intentionally shallow, hyperparameter-optimal LSTM model, which is characterized by reduced number of parameters and reduced number of layers, which limit the on-device computation requirements. Second, XGBoost models can be trained on local devices of clients, and only some light summary changes are sent to the central server, which eliminates the need to transfer full boosting trees many times. These changes collectively decrease the amount of data sent in each round and the amount of computation required in each round which directly mitigates the carbon inefficiencies observed in the previous literature without reducing the accuracy of the forecasts. Finally, the implementation approach adopted in the proposed framework, makes it a viable solution to be trained with any of the sector-wise data given in scope of the domain. 
\begin{table*}[!h]
\centering
\caption{Summary of Techniques, Application Domains, and Contributions in Forecasting and Federated Learning}
\label{tab:fl_forecasting}
\scriptsize
\renewcommand{\arraystretch}{1.2}
\begin{tabularx}{\textwidth}{|l|X|X|X|X|X|}
\hline
\textbf{Ref} & \textbf{Technique / Approach} & \textbf{Application Domain} & \textbf{Key Contribution / Finding} & \textbf{Notes / Advantages} & \textbf{Theme} \\
\hline
\cite{islam2024data} & SARIMAX + ML & Carbon emission forecasting & Integrated econometric \& ML, SARIMAX improved over ARIMA & Included external factors for precision & Forecasting \\
\hline
\cite{siami2019performance} & SARIMA vs LSTM & CO$_2$ emission prediction & LSTM better for nonlinear patterns & LSTM outperformed statistical models & Forecasting \\
\hline
\cite{liu2024online} & CNN--RNN hybrid & CO$_2$ emission forecasting & Outperformed ARIMA/SARIMAX & Captured spatio-temporal features & Forecasting \\
\hline
\cite{mustafa2024forecasting} & LSTM + FedAvg & Smart grid & Maintained accuracy while preserving privacy & Reduced networking load & FL in Energy \\
\hline
\cite{savazzi2023energy} & LSTM + FedAvg + K-means & Smart grid & Comparable to SOTA, better privacy \& training time & Added client clustering & FL in Energy \\
\hline
\cite{chen2020fedhealth} & FedHealth (Transfer Learning) & Wearable healthcare & Personalized FL via transfer learning & Improved adaptability & Personalized FL \\
\hline
\cite{liu2020carbon} & FL + Transfer Learning & Cross-domain ML & Used source labels in target domain & Better cross-domain performance & Personalized FL \\
\hline
\cite{Smith2017} & MOCHA (Multi-task FL) & Distributed ML & Tackled heterogeneity \& dropout & Reduced comm. cost & Personalized FL \\
\hline
\cite{Fallah2020} & Model Fine-Tuning (Meta-Learning) & Generic FL & Fine-tune global model locally & Balanced global \& local opt. & Personalized FL \\
\hline
\cite{khodak2019} & ARUBA (Meta-Learning) & Multi-task FL & Adaptive task similarity learning & Online convex optimization & Personalized FL \\
\hline
\cite{yang2020} & G-FML (Grouped Meta-Learning) & Multi-task FL & Grouped clients before meta-learning & Improved convergence & Personalized FL \\
\hline
\cite{zhang2022federated} & Heterogeneous Model FL & Generic FL & Allowed varying architectures & Solved model-size mismatch & Personalized FL \\
\hline
\cite{yu2021fedkd} & FedKD (Knowledge Distillation) & Generic FL & Adaptive distillation + gradient compression & Lower comm. cost & Personalized FL \\
\hline
\cite{zhu2021} & MetaFed & Cross-federation FL & Enabled knowledge sharing across FL networks & Enhanced personalization & Personalized FL \\
\hline
\cite{long2022} & Multi-center Aggregation & Generic FL & Cluster-based FL aggregation & Reduced heterogeneity effects & Personalized FL \\
\hline
\cite{yoo2021} & Federated Clustering & Healthcare & HRV-based depression severity prediction & Better accuracy with non-iid data & Personalized FL \\
\hline
\textbf{Proposed Study} & \textbf{LSTM + Attention + XGBoost + FedAvg (Transfer Learning)} & \textbf{CO$_2$ emission forecasting} & \textbf{Applied Feature Engineering with ARIMA and GARCH Features} & \textbf{Improved accuracy with non-iid data Lesser Complexity during Transfer Learning} & \textbf{Forecasting with Data Security} \\
\hline
\end{tabularx}
\end{table*}

\section{Materials and Methods}
\begin{figure*}[!t]  
\centering
\includegraphics[width=\textwidth]{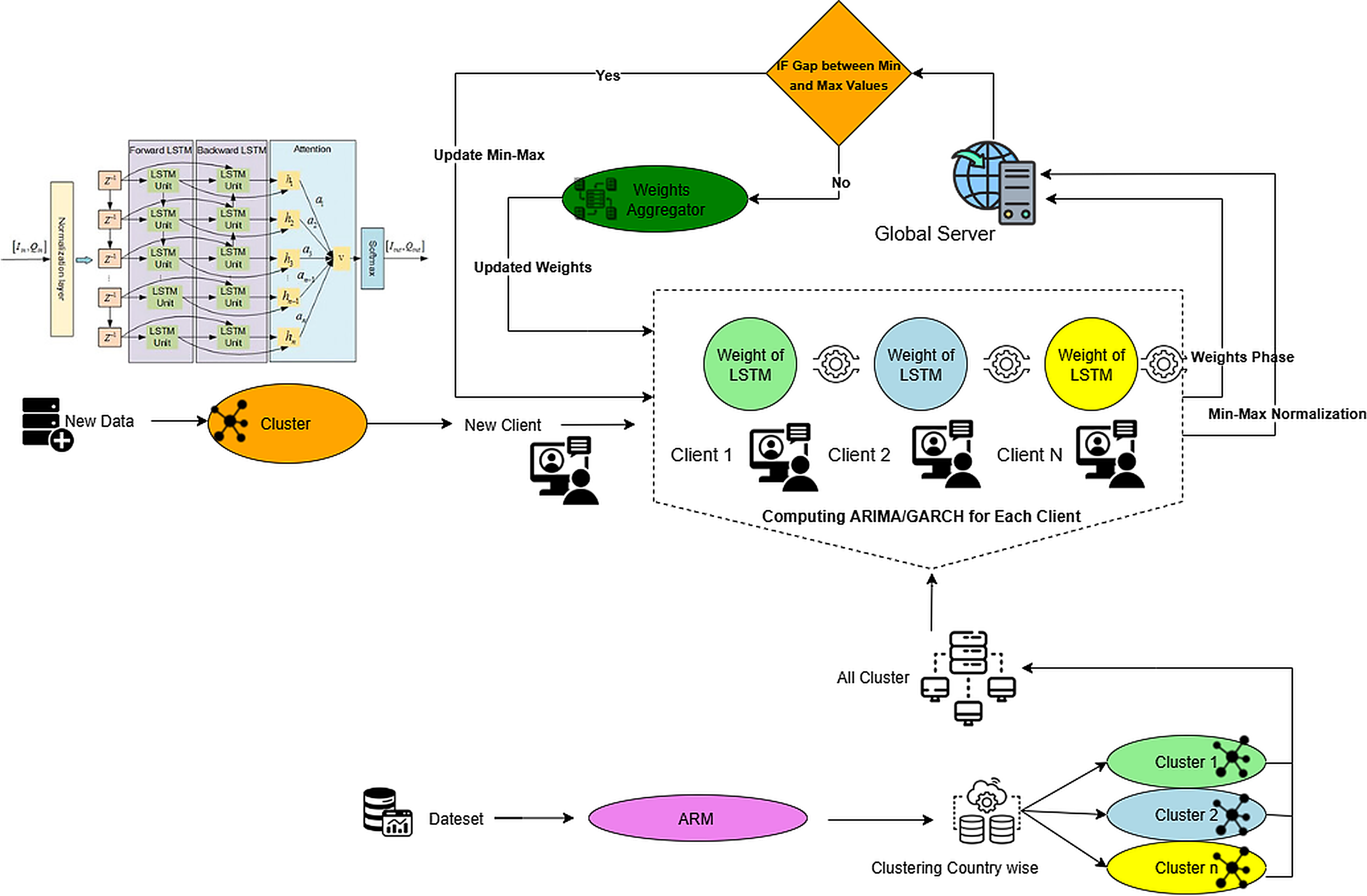} % or .png/.jpg
\caption{Proposed Framework}
\label{fig:architecture}
\end{figure*}

This research presents a federated learning framework for predicting carbon emissions through the integration of hybrid time-series modelling and deep learning techniques.  The methodology aims to integrate data from many industries across multiple countries, considering each country or region as an autonomous learning entity termed as client in the study. Moreover, the proposed study employs both statistical and neural modelling techniques within a federated framework where clients can learn through temporal features and pass on the weights to the centralized system. The centralized system collect the weights from various clients and update the learning model and forward the newly learned weight to each client (as shown in Fig.~\ref{fig:architecture}). Further, Fig.~\ref{fig:architecture} shows detailed framework diagram and flow of the data in the proposed architecture. In first step data is divided into respective regions or countries each acting as individual client. Further, ARIMA and GARCH features are computed for each client and individual training process is initiated. Then the min-max and weight values are transferred to the GlobalServer, where it checks for min-max consistency. In case of any inconsistent value it assigns a global minima and maxima and forwards it to clients to re-normalize accordingly, else it passes the weights to aggregator (FedAvg). After computing the new weights the weights are transferred back to the clients. Similarly, if a new client tries to add into the framework the same process is repeated with only training for the new client.  
% Details of Figure 1 added above. Also, it explains the procesdure of Min-Max normalization updates.

The dataset used in this study consists of carbon-emission records collected across multiple countries and emission sectors, covering the period up to 2025. Unlike previous studies \cite{cui2023federated,niu2020can}, which primarily focused on individual sectors, the proposed approach aggregates emissions from multiple sectors to obtain total carbon-emission values for forecasting. The original dataset records sector information through a single \texttt{sector} attribute, which was subsequently transformed into separate sector-specific variables representing Domestic Aviation, Ground Transport, Industry, International Aviation, Power, and Residential emissions. This preprocessing strategy enables flexible formulation of forecasting tasks by selecting either total emissions or individual sector emissions as the prediction target. In the present study, experimental evaluation is conducted using total carbon-emission data, while sector-specific forecasting remains a potential application of the proposed framework. Furthermore, the federated architecture is designed to support the incorporation of additional clients through local training and aggregation procedures without requiring data sharing among existing participants.

% It is worth noting that the dataset utilised is in agreement with the one utilised in various reference studies \cite{cui2023federated,niu2020can} of the domain, which encompasses the time period up until the year 2025.  In contrast to the previous reference studies \cite{liu2024online, cui2023federated}, which concentrated on one particular sector, the proposed approach of this study aggregates the entire emission value by summing emissions from five separate sectors.  The emissions by sector were first recorded in a single attribute designated as \texttt{sector}.  This feature was systematically divided into five distinct columns, each representing a specific sector: Domestic Aviation, Ground Transport, Industry, International Aviation, Power, and Residential. The experiments are conducted in such a manner that the model can easily be trained to forecast not only the total carbon dioxide emission but can be shifted to any of the said sectors with simple retraining process. Furthermore, the proposed framework provides the facility to add new clients with minimal training process required without affecting the other clients of the Federated Network.

\subsection{Data Preparation}
The data preprocessing incorporates temporal alignment, addressing missing values, and normalising emission levels throughout each sector. Furthermore, clustering approach is employed to enable distributed modelling, considering each nation or region in the dataset as a distinct client. This geographical segmentation facilitates country-specific modelling and endorses the federated learning protocol. Every individual client experiences noise reduction and scaling using Min-Max normalisation to guarantee consistency of input sequences for training. 
There were two key places to perform the preprocessing and normalisation of data. Firstly, on the centralized level, but as this study focuses on data privacy and performs most of the operations on the client level. Therefore, preprocessing steps has been performed on the client side. This leads to an issue of inconsistent normalized value that are tackled by passing the scaler transformer weights to the federated system. If all the weights are in range of certain standard deviation the process of preprocessing is passed and clients are sent with aknowledgement, while in reciprocal case, it is notified that one of the client has different range of data and normalization process will be repeated with increased min-max values.

\subsection{Federated LSTM based Model}
To accurately capture the complex temporal and nonlinear interactions in carbon emission time-series data and improve prediction precision, a federated learning-based framework for carbon emission forecasting has been proposed. A sliding time window approach was applied to convert sequential emission data into a supervised learning format, with previous carbon emission values functioning as inputs to forecast future values.  An enhanced Long Short-Term Memory (LSTM) network incorporating an attention mechanism has been employed to predict temporal patterns and dynamically concentrate on appropriate time steps.  The latent features obtained from the LSTM-attention model were then passed to an XGBoost regressor to enhance the prediction using gradient-boosted decision trees. The Federated Averaging (FedAvg) method \cite{nilsson2018performance, sun2022decentralized} was incorporated into the framework to provide distributed training across several geographic clients, enabling each client to cooperatively develop a global model while maintaining the confidentiality and security of its emission data. Table ~\ref{tab:normalized_stats} shows the statistics of normalized data across sectors. The dataset contains record for daily CO$_2$ emission values for each sector with respect to each country between 2019 and 2025. To have a consistency in the validation process, data of 2024 and 2025 is set aside for testing purposes for all the experiments. However, various chunks of test data are selected depending on the nature of the conducted experiment.
\addtocounter{table}{-1}
\begin{table}[H]
\centering
\scriptsize
\caption{Descriptive statistics of normalized CO$_2$ emissions across sectors}
\label{tab:normalized_stats}
\begin{tabular}{p{0.8cm}|p{0.8cm}|p{0.8cm}|p{0.8cm}|p{0.8cm}|p{0.8cm}|p{0.8cm}|p{0.8cm}}
\hline
Statistic & Domestic Aviation & Ground Transport & Industry & Intl. Aviation & Power & Resident & total\_CO$_2$ \\
\hline
Count  & 31948 & 31948 & 31948 & 31948 & 31948 & 31948 & 31948 \\
Mean   & 0.1133 & 0.1262 & 0.1221 & 0.0994 & 0.1201 & 0.0687 & 0.1204 \\
Std    & 0.2114 & 0.2231 & 0.2249 & 0.2105 & 0.2159 & 0.1361 & 0.2163 \\
Min    & 0.0000 & 0.0000 & 0.0000 & 0.0000 & 0.0000 & 0.0000 & 0.0000 \\
25\%   & 0.0056 & 0.0136 & 0.0051 & 0.0054 & 0.0042 & 0.0040 & 0.0062 \\
50\%   & 0.0212 & 0.0222 & 0.0208 & 0.0111 & 0.0356 & 0.0157 & 0.0276 \\
75\%   & 0.1310 & 0.1249 & 0.0767 & 0.0673 & 0.1008 & 0.0703 & 0.1141 \\
Max    & 1.0000 & 1.0000 & 1.0000 & 1.0000 & 1.0000 & 1.0000 & 1.0000 \\
\hline
\end{tabular}
\end{table}

\subsubsection{LSTM-Attention Model}
An LSTM (Long Short-Term Memory) model enhanced with an attention mechanism is employed to capture nonlinear temporal associations. The conventional LSTM design is characterised by the subsequent update equations:

\begin{equation}
f_t = \sigma(W_f \cdot [h_{t-1}, x_t] + b_f)
\end{equation}
\begin{equation}
i_t = \sigma(W_i \cdot [h_{t-1}, x_t] + b_i)
\end{equation}
\begin{equation}
\tilde{C}_t = \tanh(W_C \cdot [h_{t-1}, x_t] + b_C)
\end{equation}
\begin{equation}
C_t = f_t * C_{t-1} + i_t * \tilde{C}_t
\end{equation}
\begin{equation}
o_t = \sigma(W_o \cdot [h_{t-1}, x_t] + b_o)
\end{equation}
\begin{equation}
h_t = o_t * \tanh(C_t)
\end{equation}

An attention mechanism is applied on top of the LSTM outputs to selectively weigh temporal features as explained through following equations:

\begin{equation}
\alpha_t = \frac{\exp(e_t)}{\sum_{k=1}^{T} \exp(e_k)}, \quad e_t = v^\top \tanh(W_a h_t + b_a)
\end{equation}
\begin{equation}
c = \sum_{t=1}^{T} \alpha_t h_t
\end{equation}

The context vector \(c\) is then passed through a feed-forward layer and used as a high-level representation of sequential emission patterns. 
The statistical features are extracted for each client as follows.

\subsubsection{ARIMA and GARCH Models}
The ARIMA model is used to model and detrend each sector's emission series, as defined by:

\begin{equation}
y_t = c + \phi_1 y_{t-1} + \cdots + \phi_p y_{t-p} + \theta_1 \epsilon_{t-1} + \cdots + \theta_q \epsilon_{t-q} + \epsilon_t
\end{equation}

where \(\phi_i\) and \(\theta_j\) are the autoregressive and moving average coefficients, respectively. In parallel, GARCH (Generalized Autoregressive Conditional Heteroskedasticity) is employed to capture volatility:

\begin{equation}
\sigma_t^2 = \omega + \sum_{i=1}^{q} \alpha_i \epsilon_{t-i}^2 + \sum_{j=1}^{p} \beta_j \sigma_{t-j}^2
\end{equation}

These statistical components are computed for each client independently and concatenated with sector features passed to the LSTM Model. The resulting high-dimensional feature vector is input to an XGBoost regressor, defined as:

\begin{equation}
\hat{y}_i = \sum_{k=1}^{K} f_k(x_i), \quad f_k \in \mathcal{F}
\end{equation}

where \(\mathcal{F}\) denotes the space of regression trees. XGBoost optimizes the following objective:

\begin{equation}
\mathcal{L}(\phi) = \sum_{i} l(y_i, \hat{y}_i) + \sum_{k} \Omega(f_k)
\end{equation}
\begin{equation}
\Omega(f) = \gamma T + \frac{1}{2} \lambda \sum_{j=1}^{T} w_j^2
\end{equation}

\subsection{Algorithmic Designs}
Furthermore, the framework given below consists of four algorithms connected to each other to support privacy-preserving hybrid forecasting in a federated learning environment. The \textbf{Algorithm~\ref{alg:local_training}} outlines the local training process and update generation of every client. During this step, a client is provided with the existing global model parameters and preprocesses its time-series data through ARIMA and GARCH statistical tools and localized training of LSTM network to learn latent representations. An XGBoost regressor is then trained in the combined feature space on the combination of the ARIMA/GARCH descriptors plus the latent variables obtained through the use of LSTM. The difference between the client-related parameters and the global parameters is calculated so as to create a model update which is sent to the central server.  

% \begin{algorithm}[H]
% \scriptsize
% \caption{Local Training and Model Update at Client $k$}
% \label{alg:local_training}
% \begin{algorithmic}[1]
% \Require Local dataset $D_k$ for client $k$ (time-series CO$_2$ emissions + ARIMA and GARCH features); global models $\{\text{LSTM}^t, \text{XGBoost}^t\}$ received from server at round $t$
% \Ensure Model updates $\Delta \text{LSTM}_k$ and $\Delta \text{XGBoost}_k$ sent to server
% \State Receive global models $\{\text{LSTM}^t, \text{XGBoost}^t\}$ from the server
% \State Preprocess local data: normalize time-series data and construct additional features
% \State Train local LSTM model $\text{LSTM}^k$ on preprocessed data
% \State Extract hidden states or final output layer from $\text{LSTM}^k$
% \State Combine LSTM output with external features (ARIMA, GARCH, contextual features)
% \State Train local XGBoost model $\text{XGBoost}^k$ on combined feature set
% \State Compute model updates:
% \[
% \Delta \text{LSTM}_k = \text{LSTM}^t - \text{LSTM}^k, 
% \]
% \[
% \Delta \text{XGBoost}_k = \text{XGBoost}^t - \text{XGBoost}^k
% \]
% \State Send updates $\Delta \text{LSTM}_k$, $\Delta \text{XGBoost}_k$ to the server
% \end{algorithmic}
% \end{algorithm}

\begin{algorithm}[H]
\scriptsize
\caption{Local Training and Model Update at Client $k$}
\label{alg:local_training}
\begin{algorithmic}[1]

\Require Local dataset $D_k$ containing time-series CO$_2$ emissions, ARIMA, GARCH, and contextual features; global models $\{\mathrm{LSTM}^{t}, \mathrm{XGBoost}^{t}\}$ received from the server at round $t$

\Ensure Local model updates $\Delta \mathrm{LSTM}_k$ and $\Delta \mathrm{XGBoost}_k$

\State Receive global models $\{\mathrm{LSTM}^{t}, \mathrm{XGBoost}^{t}\}$ from the server

\State Normalize local time-series data and construct ARIMA, GARCH, and contextual features

\State Train local LSTM model $\mathrm{LSTM}^{k}$ using the preprocessed dataset

\State Extract temporal representations from $\mathrm{LSTM}^{k}$

\State Fuse LSTM representations with ARIMA, GARCH, and contextual features

\State Train local XGBoost model $\mathrm{XGBoost}^{k}$ using the fused feature set

\State Compute model updates:
\State \hspace{1em} $\Delta \mathrm{LSTM}_k \leftarrow \mathrm{LSTM}^{k} - \mathrm{LSTM}^{t}$
\State \hspace{1em} $\Delta \mathrm{XGBoost}_k \leftarrow \mathrm{XGBoost}^{k} - \mathrm{XGBoost}^{t}$

\State Send $\Delta \mathrm{LSTM}_k$ and $\Delta \mathrm{XGBoost}_k$ to the server

\State \Return $(\Delta \mathrm{LSTM}_k,\Delta \mathrm{XGBoost}_k)$

\end{algorithmic}
\end{algorithm}

The \textbf{Algorithm~\ref{alg:server_aggregation}} recommends the federated aggregation process that is carried out at the server. This one has aggregation of multiple client updates over weighted averaging, where weights are in proportion to the volume of data in each client. The combination of the resulting update is fed back to update the global LSTM and XGBoost model; the updated model is subsequently distributed back to all the clients involved.  

% \begin{algorithm}[H]
% \caption{Federated Learning Aggregation at Server}
% \label{alg:server_aggregation}
% \begin{algorithmic}[1]
% \scriptsize
% \Require Model updates $\{\Delta \text{LSTM}_k, \Delta \text{XGBoost}_k\}_{k=1}^K$; client weights $p_k$; global models $\{\text{LSTM}^t, \text{XGBoost}^t\}$
% \Ensure Updated global models $\{\text{LSTM}^{t+1}, \text{XGBoost}^{t+1}\}$
% \For{each round $t$}
%     \State Receive updates $\Delta \text{LSTM}_k, \Delta \text{XGBoost}_k$ from all clients
%     \State Aggregate using weighted averaging:
%     \[
%     \Delta \text{LSTM}^{t+1} = \frac{\sum_{k=1}^K p_k \Delta \text{LSTM}_k}{\sum_{k=1}^K p_k}, 
%     \]
%     \[
%     \Delta \text{XGBoost}^{t+1} = \frac{\sum_{k=1}^K p_k \Delta \text{XGBoost}_k}{\sum_{k=1}^K p_k}
%     \]
%     \State Update global models:
%     \[
%     \text{LSTM}^{t+1} = \text{LSTM}^t - \Delta \text{LSTM}^{t+1} \]
%     \[\text{XGBoost}^{t+1} = \text{XGBoost}^t - \Delta \text{XGBoost}^{t+1}
%     \]
%     \State Broadcast updated global models $\{\text{LSTM}^{t+1}, \text{XGBoost}^{t+1}\}$ to all clients
% \EndFor
% \end{algorithmic}
% \end{algorithm}

\begin{algorithm}[H]
\scriptsize
\caption{Federated Learning Aggregation at Server}
\label{alg:server_aggregation}
\begin{algorithmic}[1]

\Require Client updates $\{\Delta \mathrm{LSTM}_k,\Delta \mathrm{XGBoost}_k\}_{k=1}^{K}$; client weights $\{p_k\}_{k=1}^{K}$; global models $\{\mathrm{LSTM}^{t},\mathrm{XGBoost}^{t}\}$

\Ensure Updated global models $\{\mathrm{LSTM}^{t+1},\mathrm{XGBoost}^{t+1}\}$

\For{each communication round $t$}

    \State Receive updates $\Delta \mathrm{LSTM}_k$ and $\Delta \mathrm{XGBoost}_k$ from participating clients

    \State Aggregate client updates using weighted averaging:
    \State \hspace{1em}
    $\Delta \mathrm{LSTM}^{t+1}
    \leftarrow
    \dfrac{\sum_{k=1}^{K} p_k \Delta \mathrm{LSTM}_k}
    {\sum_{k=1}^{K} p_k}$

    \State \hspace{1em}
    $\Delta \mathrm{XGBoost}^{t+1}
    \leftarrow
    \dfrac{\sum_{k=1}^{K} p_k \Delta \mathrm{XGBoost}_k}
    {\sum_{k=1}^{K} p_k}$

    \State Update global models:
    \State \hspace{1em}
    $\mathrm{LSTM}^{t+1}
    \leftarrow
    \mathrm{LSTM}^{t}
    -
    \Delta \mathrm{LSTM}^{t+1}$

    \State \hspace{1em}
    $\mathrm{XGBoost}^{t+1}
    \leftarrow
    \mathrm{XGBoost}^{t}
    -
    \Delta \mathrm{XGBoost}^{t+1}$

    \State Broadcast updated models
    $\{\mathrm{LSTM}^{t+1},\mathrm{XGBoost}^{t+1}\}$
    to all clients

\EndFor

\State \Return $\{\mathrm{LSTM}^{t+1},\mathrm{XGBoost}^{t+1}\}$

\end{algorithmic}
\end{algorithm}

The \textbf{Algorithm~\ref{alg:prediction}} defines the prediction stage. At this stage, a set of global LSTM and XGBoost models is used to produce predictions in held out test data. The two models are summed together to give a forecast and it is estimated through the application of standard regression tools, such as coefficient of determination ($R^2$), mean squared error (MSE), root mean squared error (RMSE), mean absolute error (MAE), and mean absolute percentage error (MAPE).

\begin{algorithm}[H]
\scriptsize
\caption{Prediction and Ensemble Aggregation}
\label{alg:prediction}
\begin{algorithmic}[1]

\Require Final global models $\{\mathrm{LSTM}^{t+1}, \mathrm{XGBoost}^{t+1}\}$; test dataset $D_{\mathrm{test}}$; ensemble weight $\alpha$

\Ensure Predicted CO$_2$ emissions $\hat{y}$ and evaluation metrics $\{R^2,\mathrm{MSE},\mathrm{RMSE},\mathrm{MAE},\mathrm{MAPE}\}$

\For{each test sample in $D_{\mathrm{test}}$}

    \State Generate LSTM prediction:
    \State \hspace{1em}
    $y_{\mathrm{LSTM}}
    \leftarrow
    \mathrm{LSTM}^{t+1}(\mathrm{test\_sample})$

    \State Generate XGBoost prediction:
    \State \hspace{1em}
    $y_{\mathrm{XGB}}
    \leftarrow
    \mathrm{XGBoost}^{t+1}(\mathrm{test\_sample})$

    \State Compute ensemble prediction:
    \State \hspace{1em}
    $\hat{y}
    \leftarrow
    \alpha\, y_{\mathrm{LSTM}}
    +
    (1-\alpha)\, y_{\mathrm{XGB}}$

\EndFor

\State Compute evaluation metrics:
\State \hspace{1em}
$\{\mathrm{MSE}, \mathrm{RMSE}, \mathrm{MAE}, \mathrm{MAPE}, R^2\}$

\State \Return $\{\hat{y}, R^2, \mathrm{MSE}, \mathrm{RMSE}, \mathrm{MAE}, \mathrm{MAPE}\}$

\end{algorithmic}
\end{algorithm}

% \begin{algorithm}[H]
% \caption{Federated Averaging (FedAvg) – General Approach}
% \scriptsize
% \label{alg:fedavg}
% \begin{algorithmic}[1]
% \Require $K$ clients with local models $\{w_k^t\}_{k=1}^K$ at round $t$
% \Ensure Updated global model $w^{t+1}$
% \State Initialize global model $w^0$
% \For{each round $t = 1, 2, \dots$}
%     \For{each client $k = 1, \dots, K$}
%         \State Train on local dataset $D_k$ to obtain $w_k^t$
%     \EndFor
%     \State Aggregate client models:
%     \[
%     w^{t+1} = \sum_{k=1}^{K} \frac{n_k}{n} w_k^t
%     \]
%     \hfill \textit{where $n_k$ is client $k$'s sample size and $n = \sum_{k=1}^K n_k$}
%     \State Distribute updated global model $w^{t+1}$ to all clients
% \EndFor
% \end{algorithmic}
% \end{algorithm}

\subsection{Ablation Study}
In order to assess the effectiveness of various architectural combinations, an ablation study is performed. This study examines five model variants: (1) standalone LSTM, (2) LSTM with attention, (3) XGBoost regressor applied directly to sectoral features, (4) LSTM with attention followed by XGBoost, and (5) the final integrated model combining ARIMA and GARCH statistical features with LSTM-attention and XGBoost. 
The performance of these five experimental settings on three selected clients are summarized in Table~\ref{tab:ablation_client1} for client 1, Table~\ref{tab:ablation_client2} for client 2, and Table~\ref{tab:ablation_client3} for client 3. Also, Table~\ref{tab:hyperparams_summary} describes the hyperparameterr settings for various models used during various experiments.

The final hybrid model (LSTM + XGBoost) with ARIMA and GARCH features has been proven most effective and is integrated into a federated learning setup using the FedAvg algorithm. In this setting, each country acts as a client node with private emission data. Each client locally trains the LSTM-attention and XGBoost model on its region-specific data and shares only the model weights, not the data. These local weights are then averaged at a central server using the FedAvg strategy. Further, each time new emission data is recorded or a new country client joins the network, the corresponding local model is updated, and the global weights are re-averaged using FedAvg. This decentralized training paradigm ensures data privacy while enabling global learning efficiency. Over time, the system becomes increasingly robust to client variability and model drift, ensuring scalable and privacy-preserving carbon emission forecasting across heterogeneous regions.

\begin{table}[H]
\centering
\scriptsize
\caption{Hyperparameter Settings for All Model Variants}
\label{tab:hyperparams_summary}
\begin{tabular}{p{3.5cm}|p{4.2cm}}
\hline
\textbf{Model Variant} & \textbf{LSTM/XGBoost Hyperparameters} \\
\hline
LSTM Only & LSTM (50 units), ReLU, Dropout=0.2, Adam, 20 epochs, Dense(1) \\
LSTM + Attention & LSTM (50), Attention, Dropout=0.2, Adam, Dense(50,16,1), 20 epochs \\
XGBoost Only & XGBoost: 100 estimators, LR=0.1, default depth \\
LSTM + Attention + XGBoost & LSTM(50), Attention, Dense(50,16), Dropout=0.2, XGBoost(100 est.) \\
ARIMA + GARCH + LSTM + XGBoost & ARIMA + GARCH features + LSTM(50), Attention, Dense(50,16), Dropout=0.2, XGBoost(100 est.) \\
\hline
\end{tabular}
\end{table}

\begin{table}[H]
\centering
\scriptsize
\caption{Ablation Study for Client 1: Model Variants, Metrics, and Hyperparameters}
\label{tab:ablation_client1}
\begin{tabular}{p{5cm}|p{0.8cm}|p{0.8cm}|p{0.8cm}}
\hline
\textbf{Model Variant} & \textbf{RMSE} & \textbf{MAE} & \textbf{R\textsuperscript{2}}  \\
\hline
LSTM Only & 0.4080 & 0.1781 & 0.71 \\
LSTM + Attention & 0.1080 & 0.0781 & 0.78 \\
XGBoost Only & 0.3900 & 0.1680 & 0.65  \\
LSTM + Attention + XGBoost & 0.1331 & 0.0975 & 0.85 \\
ARIMA + GARCH + LSTM + XGBoost & 0.0826 & 0.0625 & 0.91  \\
\hline
\end{tabular}
\end{table}

\begin{table}[H]
\centering
\scriptsize
\caption{Ablation Study for Client 2: Model Variants, Metrics, and Hyperparameters}
\label{tab:ablation_client2}
\begin{tabular}{p{5cm}|p{0.8cm}|p{0.8cm}|p{0.8cm}}
\hline
\textbf{Model Variant} & \textbf{RMSE} & \textbf{MAE} & \textbf{R\textsuperscript{2}}  \\
\hline
LSTM Only & 1.4092 & 1.4578 & 0.67 \\
LSTM + Attention & 1.3320 & 1.2550 & 0.70 \\
XGBoost Only & 1.3820 & 1.3120 & 0.69  \\
LSTM + Attention + XGBoost & 1.4443 & 0.9855 & 0.74 \\
ARIMA + GARCH + LSTM + XGBoost & 1.0501 & 0.7399 & 0.79  \\
\hline
\end{tabular}
\end{table}

\begin{table}[H]
\centering
\scriptsize
\caption{Ablation Study for Client 3: Model Variants, Metrics, and Hyperparameters}
\label{tab:ablation_client3}
\begin{tabular}{p{5cm}|p{0.8cm}|p{0.8cm}|p{0.8cm}}
\hline
\textbf{Model Variant} & \textbf{RMSE} & \textbf{MAE} & \textbf{R\textsuperscript{2}}  \\
\hline
LSTM Only & 0.9882 & 0.7091 & 0.73 \\
LSTM + Attention & 0.9350 & 0.6620 & 0.76 \\
XGBoost Only & 0.9750 & 0.6830 & 0.74  \\
LSTM + Attention + XGBoost & 0.9540 & 0.7261 & 0.78 \\
ARIMA + GARCH + LSTM + XGBoost & 0.6264 & 0.4928 & 0.84  \\
\hline
\end{tabular}
\end{table}

\section{Experimental Setup and Results}
In these experiments, we have followed the standard procedures of federated learning on the Carbon Emission Dataset \cite{cui2023federated}.  The target variable has been \texttt{total\_CO$_2$} as the response and enrich features with ARIMA and GARCH descriptors. By computing ARIMA and GARCH as features makes our proposed method a generalize framework that can be used with any sector value. However, due to time constraints, we have only performed experiments with \texttt{total\_CO$_2$} values. To better understand and evaluate the proposed framework, we applied different portions of data and each execution is conducted k-folds (where k is taken as 5). The mentioned results are based on average from the k-fold experiments. In the given setup, Clients correspond to countries/regions, and further we evaluate 30\%, 50\%, and 100\% data formations and aggregate with FedAvg over 5, 9, and all clients.  

\subsection{Experimental Setup}
In the initial setup we converted sector-level time series from rows to columns and then trained forecasting models on the transformed panel. We target \texttt{total\_CO$_2$} as the regression response. Feature engineering augments each client (country/region) series with autoregressive and volatility descriptors from ARIMA/SARIMA and GARCH. Model training proceeds in a cross-silo Federated Learning (FL) setup where each country/region is a client. We consider three data formations—30\%, 50\%, and 100\% of each client’s historical series—and, for each formation, we aggregate across (i) 5 clients, (ii) 9 clients, and (iii) all clients using FedAvg. 
We compute the standard regression metrics: $R^2$, MSE, MAE, RMSE, and MAPE. For compactness, we report results in metric-specific tables per formation. Let $m_b$ denote the metric value ``Before FedAvg'' (local model) and $m_a$ denote the value ``After FedAvg (All clients)''. The improvement percentage is
\[
\text{Change}(\%) =
\begin{cases}
\frac{m_a - m_b}{m_b}\times 100, & \text{for } R^2 \\
\frac{m_b - m_a}{m_b}\times 100, & \text{for error metrics}
\end{cases}
\]
This convention reflects that larger is better for $R^2$ and smaller is better for error metrics.
\subsection{Computational Complexity Analysis}

When a \texttt{CarbonClient} object is created, the code first executes a rolling ARIMA and GARCH forecasting loop for approximately $n$ steps (after the warmup), where each step fits a time-series model to a growing prefix of the dataset. Since each fit operates on increasingly larger prefixes, the total preprocessing cost becomes quadratic, i.e., $O(n^{2})$. After preprocessing, the scaling and sliding window construction are linear, $O(n \cdot T)$, while building the Keras GRU/DNN model is $O(P)$ in terms of the number of parameters, which is negligible compared to training. Training the GRU encoder is dominated by gradient backpropagation over all sequences and epochs, yielding complexity $O(E \cdot n \cdot T \cdot U^{2})$, where $E$ is the number of epochs and $U$ the hidden units. After training, latent feature extraction is $O(n \cdot T \cdot U)$, and local XGBoost fitting adds $O(M \cdot n \cdot \log n \cdot D)$ where $M$ is the number of trees and $D$ the latent dimension. Thus, the full per-client pipeline is roughly $O(n^{2} + E\cdot n\cdot T\cdot U^{2} + M\cdot n\cdot \log n \cdot D)$. 

Updating weights on a client requires setting $P$ parameters, $O(P)$, and recomputing latent features, $O(n \cdot T \cdot U)$, so update time is $O(P + n \cdot T \cdot U)$.  
On the federated server, aggregating weights across $C$ clients is $O(C \cdot P)$, distributing them is also $O(C \cdot P)$, and fitting the global XGBoost over all pooled features costs $O(M \cdot N_{\text{total}} \cdot \log N_{\text{total}} \cdot D)$ where $N_{\text{total}}=\sum n_{i}$. Communication latency per round arises from transferring weights of size $S$ over bandwidth $BW$ with latency $L$, approximated as $S/BW + L$. Each round requires one upload and one download per client, so per round cost per client is $2\cdot(S/BW + L)$. End-to-end round time is therefore $\max_{i}$ (client upload) + $O(C \cdot P)$ aggregation + distribution + global XGBoost time.

Table \ref{tab:complexity_summary} shows the complexity records of each phase in terms of asymptotic complexity measures, where $n$ denotes the number of training samples per client (2282), $T$ is the sequence window size, $E$ is the number of training epochs, $U$ is the number of hidden units in the GRU layer, $P$ denotes the number of model parameters, $M$ is the number of trees in XGBoost, $D$ is the latent feature dimension, $C$ is the number of clients (14), $N_{\mathrm{total}}$ is the total number of samples across all clients, $S$ is the model size in bytes, $BW$ is the network bandwidth, and $L$ is the network latency.

Furthermore, Tables \ref{tab:client_training_sci} and \ref{tab:client_update_sci} present the execution times for the training and update phases of each client. These values were obtained using Python-based computations and subsequently aligned with the asymptotic complexity analysis to minimize bias. All experiments were conducted in a controlled environment on a local workstation equipped with an Intel Core i3 processor and 16 GB of RAM. The implementation was developed in Python 3.9.0 using the Visual Studio Code environment. An object-oriented design was adopted to modularize the training and update phases, enabling efficient memory management by minimizing unnecessary access to secondary storage. No external accelerators (e.g., GPUs) were used; all operations were executed on CPU-based tensor computations to ensure stability and consistency across runs. Random seeds were fixed for reproducibility, and library versions were standardized for compatibility with Python 3.9.0 to reduce software dependency discrepancies.

\begin{table}[H]
\centering
\caption{Computational complexity of the proposed federated LSTM--XGBoost framework.}
\label{tab:complexity_summary}
\scriptsize
\renewcommand{\arraystretch}{1.15}
\begin{tabular}{p{4.2cm} p{3.2cm}}
\toprule
\textbf{Operation} & \textbf{Time Complexity} \\
\midrule

Client preprocessing\\
(rolling ARIMA + GARCH fits)
& $O(n^2)$ \\

Sequence construction\\
and scaling
& $O(nT)$ \\

Model build\\
(weight allocation)
& $O(P)$ \\

LSTM/GRU training\\
($E$ epochs)
& $O(E\,n\,T\,U^2)$ \\

Latent feature extraction\\
(forward pass)
& $O(nTU)$ \\

Local XGBoost fit\\
on latent features
& $O(M\,n\,\log n\,D)$ \\

Client update\\
(set weights + recompute latents)
& $O(P+nTU)$ \\

Server weight aggregation\\
($C$ clients)
& $O(CP)$ \\

Server model distribution\\
($C$ clients)
& $O(CP)$ + communication \\

Global XGBoost training\\
(on pooled latent features)
& $O(M\,N_{\mathrm{total}}
\log N_{\mathrm{total}}\,D)$ \\

Network transfer\\
(model size $S$)
& $\dfrac{S}{BW}+L$ \\

Federated round\\
(end-to-end)
& $\max_i(T_i^{\mathrm{upload}})
+ O(CP)$ \\

\bottomrule
\end{tabular}
\end{table}

% \begin{table}[H]
% \centering
% \scriptsize
% \caption{Asymptotic (Big-O) complexity of key operations}
% \label{tab:complexity_summary}
% \begin{tabular}{p{3cm} p{5cm}}
% \hline
% \textbf{Operation} & \textbf{Time complexity (Big-O)} \\
% \hline
% Client preprocessing (rolling ARIMA + GARCH fits) & $O(n^{2})$ \\
% Sequence construction, scaling & $O(n\cdot T)$ \\
% Model build (allocate weights) & $O(P)$ \\
% LSTM/GRU training (E epochs) & $O(E \cdot n \cdot T \cdot U^{2})$ \\
% Latent feature extraction (forward pass) & $O(n \cdot T \cdot U)$ \\
% Local XGBoost fit on latent features & $O(M \cdot n \cdot \log n \cdot D)$ \\
% Client update (set weights + recompute latents) & $O(P + n \cdot T \cdot U)$ \\
% Server weight aggregation (C clients) & $O(C \cdot P)$ \\
% Server distribute weights (C clients) & $O(C \cdot P)$ + network sends \\
% Global XGBoost on pooled latents (all clients) & $O(M \cdot N_{\text{total}} \cdot \log N_{\text{total}} \cdot D)$ \\
% Network transfer of weights (size $S$) & $\dfrac{S}{BW} + L$ per transfer \\
% Federated round end-to-end & $\max_{i}$(upload) + $O(C \cdot P)$ + distribution + global XGBoost \\
% \hline
% \end{tabular}
% \end{table}

\begin{table}[H]
\centering
\scriptsize
\caption{Per-client training execution times in scientific notation (ms) with normalized exponent $10^5$.}
\label{tab:client_training_sci}
\begin{tabular}{p{2cm}|p{1.5cm}|p{1.5cm}|p{1.5cm}}
\hline
\textbf{Client} & \textbf{Train 30\% ($\times 10^5$ ms)} & \textbf{Train 50\% ($\times 10^5$ ms)} & \textbf{Train 100\% ($\times 10^5$ ms)} \\
\hline
Brazil          & 1.68 & 6.61 & 10.27 \\
China           & 1.63 & 6.41 & 10.00 \\
EU27 \& UK      & 1.56 & 6.17 & 9.59  \\
France          & 1.76 & 6.93 & 10.76 \\
Germany         & 1.52 & 5.98 & 9.29  \\
India           & 1.60 & 6.30 & 9.78  \\
Italy           & 1.55 & 6.12 & 9.49  \\
Japan           & 1.64 & 6.49 & 10.08 \\
ROW             & 1.47 & 5.80 & 9.00  \\
Russia          & 1.72 & 6.80 & 10.57 \\
Spain           & 1.58 & 6.24 & 9.68  \\
United Kingdom  & 1.61 & 6.36 & 9.88  \\
United States   & 1.69 & 6.68 & 10.38 \\
WORLD           & 1.47 & 5.80 & 9.00  \\
\hline
\end{tabular}
\end{table}

\begin{table}[H]
\centering
\scriptsize 
\caption{Per-client update execution times in scientific notation (ms) with normalized exponent $10^5$.}
\label{tab:client_update_sci}
\begin{tabular}{p{2cm}|p{1.5cm}|p{1.5cm}|p{1.5cm}}
\hline
\textbf{Client} & \textbf{Update 30\% ($\times 10^5$ ms)} & \textbf{Update 50\% ($\times 10^5$ ms)} & \textbf{Update 100\% ($\times 10^5$ ms)} \\
\hline
Brazil          & 0.54 & 1.41 & 1.76 \\
China           & 0.53 & 1.38 & 1.72 \\
EU27 \& UK      & 0.54 & 1.39 & 1.74 \\
France          & 0.55 & 1.42 & 1.78 \\
Germany         & 0.53 & 1.36 & 1.71 \\
India           & 0.54 & 1.39 & 1.74 \\
Italy           & 0.52 & 1.35 & 1.68 \\
Japan           & 0.55 & 1.43 & 1.79 \\
ROW             & 0.51 & 1.32 & 1.65 \\
Russia          & 0.56 & 1.45 & 1.81 \\
Spain           & 0.53 & 1.38 & 1.72 \\
United Kingdom  & 0.54 & 1.39 & 1.74 \\
United States   & 0.55 & 1.42 & 1.78 \\
WORLD           & 0.52 & 1.34 & 1.67 \\
\hline
\end{tabular}
\end{table}

\subsection{Results: 30\% Data}
Tables~\ref{tab:r2_30}--\ref{tab:rmse_30} summarize client-level performance at 30\% data. At this low-data regime, FedAvg typically improves over local baselines, and performance increases as we go from 5 to 9 to all clients, consistent with stronger cross-client signal sharing.
A closer look at the metrics highlights several important patterns. First, R\textsuperscript{2} values often increase after FedAvg, as in Brazil, France, Spain, and the United Kingdom, suggesting that aggregation helps the model capture more variance in the target signal. However, in some cases (e.g., China, ROW, United States, WORLD), R\textsuperscript{2} decreases slightly. While this may look like a negative outcome, it can be interpreted positively: local models with limited data may overfit to idiosyncratic client patterns, artificially inflating R\textsuperscript{2}, whereas FedAvg tends to stabilize and regularize, yielding a model that generalizes better across clients. Thus, small drops in R\textsuperscript{2} should be considered in conjunction with error-based metrics. However, examining MSE, RMSE, MAE, and MAPE together provides deeper insight into generalization quality. For clients like Brazil, France, and Japan, improvements in R\textsuperscript{2} are aligned with reductions in all error metrics, which strongly indicates both better fit and reduced prediction error after FedAvg. In contrast, for clients such as China, ROW, and WORLD, R\textsuperscript{2} declines are accompanied by increases in MSE and RMSE, implying that more global data sharing alone may not suffice and that richer or more balanced training data may be needed to improve generalization. Interestingly, for India and Italy, changes in R\textsuperscript{2} are small, but reductions in MAE and MAPE suggest error reduction despite modest variance capture improvements, pointing to subtle benefits of aggregation even when R\textsuperscript{2} remains stable. These results emphasize that FedAvg introduces a trade-off that is it reduces local overfitting and encourages generalization across clients, but the quality of that generalization depends on both the heterogeneity of client distributions and the error metrics considered. When R\textsuperscript{2} decreases but error metrics improve, the model is likely generalizing better by avoiding variance capture of local noise. On the other hand, when both R\textsuperscript{2} and error metrics move negatively, it suggests underfitting and the need for more representative or larger data samples. 
% \vspace{-48pt}
\begin{table}[H]  
    \centering
    \scriptsize   
    \caption{R2 results (30\% data) across clients }
    \label{tab:r2_30}
    \begin{tabular}{p{1.5cm}|p{1cm}|p{1cm}|p{1cm}|p{1cm}|p{1cm}}
    \hline
    Client & Before FedAvg & After FedAvg (5) & After FedAvg (9) & After FedAvg (All) & Change \% \\
    \hline
    Brazil & 0.402 & 0.418 & 0.494 & 0.504 & 25.287 \\
China & 0.523 & 0.526 & 0.505 & 0.486 & -7.016 \\
EU27 \& UK & 0.439 & 0.450 & 0.465 & 0.476 & 8.514 \\
France & 0.374 & 0.404 & 0.463 & 0.492 & 31.524 \\
Germany & 0.476 & 0.474 & 0.476 & 0.476 & 0.003 \\
India & 0.371 & - & 0.188 & 0.375 & 1.205 \\
Italy & 0.337 & - & 0.409 & 0.391 & 16.085 \\
Japan & 0.621 & - & 0.693 & 0.683 & 10.016 \\
ROW & 0.333 & - & 0.326 & 0.300 & -9.915 \\
Russia & 0.900 & - & - & 0.884 & -1.854 \\
Spain & 0.118 & - & - & 0.225 & 90.651 \\
United Kingdom & 0.278 & - & - & 0.351 & 26.193 \\
United States & 0.304 & - & - & 0.275 & -9.532 \\
WORLD & 0.516 & - & - & 0.464 & -9.986 \\
    \hline
    \end{tabular}
\end{table}
\vspace{-0.8 em} % or -0.8em, adjust as needed
\begin{table}[H]  
    \centering
    \scriptsize   
    \caption{MSE results (30\% data) across clients }
    \label{tab:mse_30}
    \begin{tabular}{p{1.5cm}|p{1cm}|p{1cm}|p{1cm}|p{1cm}|p{1cm}}
    \hline
    Client & Before Fed & After FedAvg (5) & After FedAvg (9) & After FedAvg (All) & Change \% \\
    \hline
    Brazil & 0.014 & 0.014 & 0.012 & 0.012 & -17.012 \\
China & 2.706 & 2.689 & 2.804 & 2.913 & 7.684 \\
EU27 \& UK & 1.247 & 1.222 & 1.190 & 1.164 & -6.660 \\
France & 0.016 & 0.015 & 0.014 & 0.013 & -18.868 \\
Germany & 0.090 & 0.090 & 0.090 & 0.090 & -0.003 \\
India & 0.318 & - & 0.410 & 0.316 & -0.710 \\
Italy & 0.018 & - & 0.016 & 0.017 & -8.160 \\
Japan & 0.070 & - & 0.056 & 0.058 & -16.408 \\
ROW & 1.524 & - & 1.542 & 1.599 & 4.960 \\
Russia & 0.050 & - & - & 0.058 & 16.734 \\
Spain & 0.009 & - & - & 0.008 & -12.137 \\
United Kingdom & 0.021 & - & - & 0.019 & -10.095 \\
United States & 1.239 & - & - & 1.291 & 4.159 \\
WORLD & 10.559 & - & - & 12.428 & 10.639 \\
    \hline
    \end{tabular}
    \end{table}
\begin{table}[H]  
    \centering
    \scriptsize   
    \caption{MAE results (30\% data) across clients }
    \label{tab:mae_30}
    \begin{tabular}{p{1.5cm}|p{1cm}|p{1cm}|p{1cm}|p{1cm}|p{1cm}}
    \hline
    Client & Before Fed & After FedAvg (5) & After FedAvg (9) & After FedAvg (All) & Change \% \\
    \hline
   Brazil & 0.093 & 0.093 & 0.085 & 0.084 & -9.301 \\
China & 1.006 & 1.025 & 1.070 & 1.007 & 0.140 \\
EU27 \& UK & 0.900 & 0.896 & 0.897 & 0.897 & -0.330 \\
France & 0.099 & 0.091 & 0.087 & 0.086 & -13.333 \\
Germany & 0.238 & 0.242 & 0.246 & 0.243 & 2.128 \\
India & 0.444 & - & 0.452 & 0.376 & -15.330 \\
Italy & 0.109 & - & 0.104 & 0.106 & -2.800 \\
Japan & 0.209 & - & 0.195 & 0.194 & -7.179 \\
ROW & 1.026 & - & 1.024 & 1.047 & 2.095 \\
Russia & 0.188 & - & - & 0.198 & 5.777 \\
Spain & 0.079 & - & - & 0.072 & -8.508 \\
United Kingdom & 0.115 & - & - & 0.112 & -2.336 \\
United States & 0.868 & - & - & 0.888 & 2.276 \\
WORLD & 3.499 & - & - & 3.627 & 3.645 \\
    \hline
    \end{tabular}
    \end{table}
\begin{table}[H]  
    \centering
    \scriptsize   
    \caption{MAPE results (30\% data) across clients }
    \label{tab:mape_30}
    \begin{tabular}{p{1.5cm}|p{1cm}|p{1cm}|p{1cm}|p{1cm}|p{1cm}}
    \hline
    Client & Before Fed & After FedAvg (5) & After FedAvg (9) & After FedAvg (All) & Change \% \\
    \hline
Brazil & 0.088 & 0.088 & 0.079 & 0.079 & -10.482 \\
China & 0.032 & 0.033 & 0.035 & 0.033 & 1.320 \\
EU27 \& UK & 0.121 & 0.121 & 0.121 & 0.121 & -0.257 \\
France & 0.128 & 0.119 & 0.111 & 0.110 & -13.836 \\
Germany & 0.159 & 0.165 & 0.166 & 0.166 & 4.242 \\
India & 0.053 & - & 0.052 & 0.044 & -17.454 \\
Italy & 0.147 & - & 0.141 & 0.144 & -1.734 \\
Japan & 0.076 & - & 0.071 & 0.071 & -7.533 \\
ROW & 0.034 & - & 0.034 & 0.035 & 1.524 \\
Russia & 0.042 & - & - & 0.044 & 5.102 \\
Spain & 0.130 & - & - & 0.122 & -6.741 \\
United Kingdom & 0.129 & - & - & 0.126 & -2.272 \\
United States & 0.063 & - & - & 0.065 & 2.976 \\
WORLD & 0.035 & - & - & 0.036 & 3.048 \\
    \hline
    \end{tabular}
    \end{table}
\begin{table}[H]  
    \centering
    \scriptsize   
    \caption{RMSE results (30\% data) across clients }
    \label{tab:rmse_30}
    \begin{tabular}{p{1.5cm}|p{1cm}|p{1cm}|p{1cm}|p{1cm}|p{1cm}}
    \hline
    Client & Before Fed & After FedAvg (5) & After FedAvg (9) & After FedAvg (All) & Change \% \\
    \hline
    Brazil & 0.120 & 0.118 & 0.110 & 0.109 & -8.902 \\
China & 1.645 & 1.640 & 1.675 & 1.707 & 3.771 \\
EU27 \& UK & 1.117 & 1.105 & 1.091 & 1.079 & -3.387 \\
France & 0.126 & 0.123 & 0.117 & 0.114 & -9.927 \\
Germany & 0.300 & 0.300 & 0.300 & 0.300 & -0.001 \\
India & 0.564 & - & 0.641 & 0.562 & -0.356 \\
Italy & 0.136 & - & 0.128 & 0.130 & -4.167 \\
Japan & 0.264 & - & 0.237 & 0.241 & -8.571 \\
ROW & 1.234 & - & 1.242 & 1.265 & 2.450 \\
Russia & 0.224 & - & - & 0.242 & 8.044 \\
Spain & 0.097 & - & - & 0.091 & -6.265 \\
United Kingdom & 0.144 & - & - & 0.136 & -5.182 \\
United States & 1.113 & - & - & 1.136 & 2.058 \\
WORLD & 4.190 & - & - & 4.408 & 5.185 \\
    \hline
    \end{tabular}
    \end{table}

\subsection{Results: 50\% Data}
Tables~\ref{tab:r2_50}--\ref{tab:rmse_50} present the performance outcomes when 50\% of the data is used for training. Compared with the 30\% setting, overall patterns strengthen: $R^2$ generally increases for more countries, and error measures (MSE, RMSE, MAE, and MAPE) exhibit more consistent reductions, confirming that additional data improves the stability and generalization of FedAvg.

Looking first at $R^2$, several clients such as France ($+10.4\%$), India ($+9.4\%$), WORLD ($+5.4\%$), and Germany ($+2.4\%$) exhibit clear improvements under FedAvg relative to their local baselines. This indicates that aggregation across clients allows the model to capture more variance in the CO\textsubscript{2} emissions signal. Japan and the United States also show modest gains in $R^2$, while countries like China ($-7.9\%$), Brazil ($-4.0\%$), and EU27 \& UK ($-2.6\%$) experience slight declines, suggesting that in heterogeneous settings, global averaging may override some local variance capture. Nevertheless, even in these cases, error metrics often show complementary improvements, pointing to better generalization.

Error-based metrics reinforce these observations. France and Germany achieve simultaneous improvements across MSE, RMSE, MAE, and MAPE, showing that FedAvg both increases variance explanation and reduces prediction errors. India benefits strongly as well, with MSE decreasing by 15.2\% and RMSE by 7.9\%, despite a minor $R^2$ dip at intermediate client aggregation levels, underscoring the robustness gained from more data. Japan shows a similar pattern with reductions in MSE ($-5.1\%$) and RMSE ($-2.6\%$), supporting FedAvg’s stabilizing effect. On contrary, China, Russia, and ROW reveal deterioration in both $R^2$ and error metrics, indicating persistent challenges in generalization for clients with highly distinct or large-scale distributions.

At the aggregate level, WORLD results illustrate the benefit of scaling data: $R^2$ rises from $0.598$ to $0.630$, while MSE and RMSE both decrease substantially (by $-8.0\%$ and $-4.1\%$, respectively). This global consistency was weaker at 30\%, confirming that FedAvg’s advantage grows with more representative data.

When comparing across data regimes, the 50\% setting shows a clearer balance between reducing overfitting and enhancing generalization compared to the 30\% case. While 30\% data highlighted instances where $R^2$ dropped despite error improvements, the 50\% results demonstrate that with more data, both $R^2$ and error metrics improve simultaneously for a larger set of clients. This confirms that FedAvg benefits strongly from richer local data availability, reducing the bias–variance trade-off observed at lower data volumes.

\begin{table}[H]  
    \centering
    \scriptsize   
    \caption{R2 results (50\% data) across clients }
    \label{tab:r2_50}
    \begin{tabular}{p{1.5cm}|p{1cm}|p{1cm}|p{1cm}|p{1cm}|p{1cm}}
    \hline
    Client & Before Fed & After FedAvg (5) & After FedAvg (9) & After FedAvg (All) & Change \% \\
    \hline
    Brazil & 0.571 & 0.576 & 0.564 & 0.548 & -3.995 \\
China & 0.743 & 0.649 & 0.661 & 0.685 & -7.851 \\
EU27 \& UK & 0.587 & 0.582 & 0.562 & 0.572 & -2.563 \\
France & 0.491 & 0.517 & 0.539 & 0.542 & 10.403 \\
Germany & 0.560 & 0.586 & 0.569 & 0.573 & 2.361 \\
India & 0.618 & - & 0.596 & 0.676 & 9.386 \\
Italy & 0.485 & - & 0.484 & 0.487 & 0.487 \\
Japan & 0.710 & - & 0.707 & 0.725 & 2.078 \\
ROW & 0.339 & - & 0.356 & 0.343 & 1.394 \\
Russia & 0.942 & - & - & 0.931 & -1.226 \\
Spain & 0.254 & - & - & 0.248 & -2.254 \\
United Kingdom & 0.423 & - & - & 0.418 & -0.968 \\
United States & 0.410 & - & - & 0.416 & 1.413 \\
WORLD & 0.598 & - & - & 0.630 & 5.354 \\
    \hline
    \end{tabular}
    \end{table}
\begin{table}[H]  
    \centering
    \scriptsize   
    \caption{MSE results (50\% data) across clients }
    \label{tab:mse_50}
    \begin{tabular}{p{1.5cm}|p{1cm}|p{1cm}|p{1cm}|p{1cm}|p{1cm}}
    \hline
    Client & Before Fed & After FedAvg (5) & After FedAvg (9) & After FedAvg (All) & Change \% \\
    \hline
 Brazil & 0.010 & 0.010 & 0.011 & 0.011 & 5.310 \\
China & 1.443 & 1.974 & 1.906 & 1.772 & 22.744 \\
EU27 \& UK & 0.918 & 0.930 & 0.975 & 0.951 & 3.648 \\
France & 0.014 & 0.013 & 0.012 & 0.012 & -10.043 \\
Germany & 0.074 & 0.070 & 0.072 & 0.072 & -3.007 \\
India & 0.173 & - & 0.183 & 0.147 & -15.164 \\
Italy & 0.014 & - & 0.014 & 0.014 & -0.458 \\
Japan & 0.058 & - & 0.058 & 0.055 & -5.089 \\
ROW & 1.496 & - & 1.456 & 1.485 & -0.714 \\
Russia & 0.029 & - & - & 0.034 & 20.080 \\
Spain & 0.009 & - & - & 0.009 & 0.766 \\
United Kingdom & 0.017 & - & - & 0.018 & 0.708 \\
United States & 1.043 & - & - & 1.033 & -0.981 \\
WORLD & 14.086 & - & - & 12.964 & -7.962 \\
    \hline
    \end{tabular}
    \end{table}
\begin{table}[H]  
    \centering
    \scriptsize   
    \caption{MAE results (50\% data) across clients }
    \label{tab:mae_50}
    \begin{tabular}{p{1.5cm}|p{1cm}|p{1cm}|p{1cm}|p{1cm}|p{1cm}}
    \hline
    Client & Before Fed & After FedAvg (5) & After FedAvg (9) & After FedAvg (All) & Change \% \\
    \hline
 Brazil & 0.077 & 0.079 & 0.080 & 0.082 & 5.895 \\
China & 0.734 & 0.869 & 0.861 & 0.849 & 15.633 \\
EU27 \& UK & 0.752 & 0.767 & 0.788 & 0.773 & 2.903 \\
France & 0.087 & 0.084 & 0.084 & 0.083 & -5.362 \\
Germany & 0.214 & 0.206 & 0.207 & 0.207 & -3.106 \\
India & 0.322 & - & 0.326 & 0.302 & -6.156 \\
Italy & 0.093 & - & 0.094 & 0.094 & 1.287 \\
Japan & 0.190 & - & 0.190 & 0.184 & -2.967 \\
ROW & 0.993 & - & 1.001 & 1.021 & 2.821 \\
Russia & 0.135 & - & - & 0.149 & 10.988 \\
Spain & 0.073 & - & - & 0.075 & 2.406 \\
United Kingdom & 0.107 & - & - & 0.106 & -1.741 \\
United States & 0.782 & - & - & 0.794 & 1.583 \\
WORLD & 2.971 & - & - & 2.951 & -0.681 \\
    \hline
    \end{tabular}
    \end{table}
\begin{table}[H]  
    \centering
    \scriptsize   
    \caption{MAPE results (50\% data) across clients }
    \label{tab:mape_50}
    \begin{tabular}{p{1.5cm}|p{1cm}|p{1cm}|p{1cm}|p{1cm}|p{1cm}}
    \hline
    Client & Before Fed & After FedAvg (5) & After FedAvg (9) & After FedAvg (All) & Change \% \\
    \hline
Brazil & 0.073 & 0.073 & 0.074 & 0.076 & 4.713 \\
China & 0.024 & 0.028 & 0.028 & 0.028 & 16.209 \\
EU27 \& UK & 0.100 & 0.102 & 0.105 & 0.104 & 3.369 \\
France & 0.111 & 0.106 & 0.107 & 0.105 & -5.260 \\
Germany & 0.144 & 0.139 & 0.141 & 0.141 & -1.823 \\
India & 0.038 & - & 0.038 & 0.036 & -6.368 \\
Italy & 0.125 & - & 0.127 & 0.127 & 1.498 \\
Japan & 0.070 & - & 0.070 & 0.068 & -2.939 \\
ROW & 0.033 & - & 0.033 & 0.034 & 2.384 \\
Russia & 0.030 & - & - & 0.033 & 9.595 \\
Spain & 0.118 & - & - & 0.122 & 3.718 \\
United Kingdom & 0.120 & - & - & 0.118 & -1.684 \\
United States & 0.056 & - & - & 0.057 & 2.482 \\
WORLD & 0.030 & - & - & 0.030 & -0.907 \\
    \hline
    \end{tabular}
    \end{table}
\begin{table}[H]  
    \centering
    \scriptsize   
    \caption{RMSE results (50\% data) across clients }
    \label{tab:rmse_50}
    \begin{tabular}{p{1.5cm}|p{1cm}|p{1cm}|p{1cm}|p{1cm}|p{1cm}}
    \hline
    Client & Before Fed & After FedAvg (5) & After FedAvg (9) & After FedAvg (All) & Change \% \\
    \hline
Brazil & 0.102 & 0.101 & 0.103 & 0.104 & 2.621 \\
China & 1.201 & 1.405 & 1.381 & 1.331 & 10.790 \\
EU27 \& UK & 0.958 & 0.964 & 0.987 & 0.975 & 1.808 \\
France & 0.116 & 0.113 & 0.111 & 0.110 & -5.155 \\
Germany & 0.272 & 0.264 & 0.269 & 0.268 & -1.515 \\
India & 0.416 & - & 0.428 & 0.383 & -7.894 \\
Italy & 0.119 & - & 0.119 & 0.119 & -0.229 \\
Japan & 0.240 & - & 0.241 & 0.234 & -2.578 \\
ROW & 1.223 & - & 1.207 & 1.219 & -0.358 \\
Russia & 0.169 & - & - & 0.186 & 9.581 \\
Spain & 0.092 & - & - & 0.093 & 0.382 \\
United Kingdom & 0.132 & - & - & 0.133 & 0.354 \\
United States & 1.021 & - & - & 1.016 & -0.492 \\
WORLD & 3.753 & - & - & 3.601 & -4.064 \\
    \hline
    \end{tabular}
    \end{table}
    
\subsection{Results: 100\% Data}
Tables~\ref{tab:r2_100}--\ref{tab:rmse_100} present the 100\% data results. Compared to the 50\% case, performance patterns shift in notable ways. At the individual client level, $R^2$ values are generally high before federation (e.g., China 0.893, Russia 0.965, ROW 0.795), but many clients show deterioration once full aggregation is applied, particularly for Brazil, EU27 \& UK, Italy, ROW, and Spain. For example, Brazil’s $R^2$ drops from 0.719 locally to 0.616 after full FedAvg, while ROW declines sharply from 0.795 to 0.529. Similar losses are mirrored in error metrics: MSE, MAE, MAPE, and RMSE all increase for several regions when moving from local to global models. In contrast, a few countries (e.g., Germany and Japan) retain relatively stable performance across FedAvg configurations, with only marginal fluctuations.  

An important point to note is that although $R^2$ values decline for some clients, the visual inspection of fitted curves shows that the global model still tracks emission patterns very closely. This seeming paradox indicates that federated learning is not only suitable to the data, but also eliminating client-specific variations that can serve as noise. Given the statement, it can be deduce that the global model sacrifices some local accuracy in exchange for a more generalized and stable representation across all clients. This robustness is a strength of the proposed framework, as it highlights the ability of FL to balance diverse regional dynamics while still producing reliable overall forecasts. 

Technically, this effect is achieved by the fact that FedAvg places greater weights on clients with large and balanced sample sizes. This effect in a way is capable of averaging away variations in smaller or more volatile regions, thus provide a trade-off between region wise accuracy and global forecasting. Furthermore, the feature engineering approach based on construction of ARIMA and GARCH descriptors guarantees that the global model builds the similar temporal as well as volatility structures which are also seen in the fitted curves despite the statistical measures that may seem weaker. Therefore, the reduction in the values of the $R^2$ should be seen as a trade-off due to heterogeneous data distribution, but not loss of forecasting power. As a conclusive remarks to describe the achievements of this work, it can be materialized that distributional heterogeneity across clients appears to penalize certain regions more heavily under full aggregation. Still, the multi-client setups (5, 9, and all) show a stabilizing effect in terms of prediction smoothness compared to smaller data fractions, even if absolute metrics do not always improve.  
\begin{table}[H]  
    \centering
    \scriptsize   
    \caption{R2 results (100\% data) across clients}
    \label{tab:r2_100}
    \begin{tabular}{p{1.5cm}|p{1cm}|p{1cm}|p{1cm}|p{1cm}|p{1cm}}
    \hline
    Client & Before Fed & After FedAvg (5) & After FedAvg (9) & After FedAvg (All) & Change \% \\
    \hline
Brazil & 0.719 & 0.674 & 0.654 & 0.616 & -14.325 \\
China & 0.893 & 0.847 & 0.845 & 0.858 & -3.877 \\
EU27 \& UK & 0.842 & 0.793 & 0.785 & 0.754 & -10.436 \\
France & 0.744 & 0.724 & 0.744 & 0.730 & -1.934 \\
Germany & 0.777 & 0.778 & 0.780 & 0.760 & -2.206 \\
India & 0.813 & - & 0.791 & 0.783 & -3.651 \\
Italy & 0.791 & - & 0.688 & 0.648 & -18.108 \\
Japan & 0.839 & - & 0.822 & 0.826 & -1.566 \\
ROW & 0.795 & - & 0.631 & 0.529 & -33.431 \\
Russia & 0.965 & - & - & 0.941 & -2.410 \\
Spain & 0.667 & - & - & 0.496 & -25.692 \\
United Kingdom & 0.628 & - & - & 0.596 & -5.028 \\
United States & 0.747 & - & - & 0.693 & -7.268 \\
WORLD & 0.888 & - & - & 0.788 & -11.332 \\
    \hline
    \end{tabular}
    \end{table}
\begin{table}[H]  
    \centering
    \scriptsize   
    \caption{MSE results (100\% data) across clients}
    \label{tab:mse_100}
    \begin{tabular}{p{1.5cm}|p{1cm}|p{1cm}|p{1cm}|p{1cm}|p{1cm}}
    \hline
    Client & Before Fed & After FedAvg (5) & After FedAvg (9) & After FedAvg (All) & Change \% \\
    \hline
Brazil & 0.007 & 0.008 & 0.008 & 0.009 & 36.623 \\
China & 0.648 & 0.923 & 0.937 & 0.857 & 32.349 \\
EU27 \& UK & 0.398 & 0.523 & 0.542 & 0.619 & 55.623 \\
France & 0.008 & 0.009 & 0.008 & 0.008 & 5.618 \\
Germany & 0.044 & 0.044 & 0.043 & 0.047 & 7.704 \\
India & 0.090 & - & 0.100 & 0.104 & 15.882 \\
Italy & 0.006 & - & 0.009 & 0.011 & 68.439 \\
Japan & 0.035 & - & 0.039 & 0.038 & 8.170 \\
ROW & 0.509 & - & 0.913 & 1.167 & 129.360 \\
Russia & 0.018 & - & - & 0.029 & 65.850 \\
Spain & 0.004 & - & - & 0.006 & 51.508 \\
United Kingdom & 0.013 & - & - & 0.014 & 8.470 \\
United States & 0.488 & - & - & 0.593 & 21.463 \\
WORLD & 4.380 & - & - & 8.333 & 90.282 \\
    \hline
    \end{tabular}
    \end{table}
\begin{table}[H]  
    \centering
    \scriptsize   
    \caption{MAE results (100\% data) across clients}
    \label{tab:mae_100}
    \begin{tabular}{p{1.5cm}|p{1cm}|p{1cm}|p{1cm}|p{1cm}|p{1cm}}
    \hline
    Client & Before Fed & After FedAvg (5) & After FedAvg (9) & After FedAvg (All) & Change \% \\
    \hline
 Brazil & 0.059 & 0.067 & 0.068 & 0.074 & 27.075 \\
China & 0.557 & 0.653 & 0.649 & 0.655 & 17.509 \\
EU27 \& UK & 0.472 & 0.570 & 0.592 & 0.638 & 35.211 \\
France & 0.067 & 0.069 & 0.068 & 0.071 & 5.030 \\
Germany & 0.165 & 0.166 & 0.167 & 0.174 & 5.692 \\
India & 0.226 & - & 0.231 & 0.262 & 15.966 \\
Italy & 0.058 & - & 0.076 & 0.082 & 42.924 \\
Japan & 0.151 & - & 0.157 & 0.155 & 2.568 \\
ROW & 0.518 & - & 0.753 & 0.876 & 69.131 \\
Russia & 0.102 & - & - & 0.136 & 33.026 \\
Spain & 0.049 & - & - & 0.063 & 28.841 \\
United Kingdom & 0.092 & - & - & 0.095 & 3.161 \\
United States & 0.543 & - & - & 0.619 & 14.028 \\
WORLD & 1.525 & - & - & 2.350 & 54.113 \\
    \hline
    \end{tabular}
    \end{table}
\begin{table}[H]  
    \centering
    \scriptsize   
    \caption{MAPE results (100\% data) across clients}
    \label{tab:mape_100}
    \begin{tabular}{p{1.5cm}|p{1cm}|p{1cm}|p{1cm}|p{1cm}|p{1cm}}
    \hline
    Client & Before Fed & After FedAvg (5) & After FedAvg (9) & After FedAvg (All) & Change \% \\
    \hline
Brazil & 0.056 & 0.064 & 0.065 & 0.070 & 25.052 \\
China & 0.018 & 0.021 & 0.021 & 0.021 & 18.114 \\
EU27 \& UK & 0.059 & 0.074 & 0.076 & 0.083 & 38.889 \\
France & 0.085 & 0.087 & 0.084 & 0.089 & 4.809 \\
Germany & 0.106 & 0.107 & 0.107 & 0.113 & 5.999 \\
India & 0.027 & - & 0.027 & 0.031 & 15.022 \\
Italy & 0.072 & - & 0.099 & 0.109 & 50.624 \\
Japan & 0.055 & - & 0.058 & 0.057 & 3.815 \\
ROW & 0.017 & - & 0.025 & 0.029 & 69.528 \\
Russia & 0.022 & - & - & 0.029 & 30.324 \\
Spain & 0.077 & - & - & 0.099 & 28.925 \\
United Kingdom & 0.100 & - & - & 0.104 & 3.317 \\
United States & 0.039 & - & - & 0.045 & 15.049 \\
WORLD & 0.015 & - & - & 0.023 & 54.942 \\
    \hline
    \end{tabular}
    \end{table}
\begin{table}[H]  
    \centering
    \scriptsize   
    \caption{RMSE results (100\% data) across clients}
    \label{tab:rmse_100}
    \begin{tabular}{p{1.5cm}|p{1cm}|p{1cm}|p{1cm}|p{1cm}|p{1cm}}
    \hline
    Client & Before Fed & After FedAvg (5) & After FedAvg (9) & After FedAvg (All) & Change \% \\
    \hline
 Brazil & 0.081 & 0.087 & 0.090 & 0.095 & 16.886 \\
China & 0.805 & 0.961 & 0.968 & 0.926 & 15.043 \\
EU27 \& UK & 0.631 & 0.723 & 0.736 & 0.787 & 24.749 \\
France & 0.089 & 0.093 & 0.089 & 0.092 & 2.771 \\
Germany & 0.209 & 0.209 & 0.208 & 0.217 & 3.780 \\
India & 0.299 & - & 0.316 & 0.322 & 7.649 \\
Italy & 0.079 & - & 0.097 & 0.103 & 29.784 \\
Japan & 0.188 & - & 0.197 & 0.195 & 4.005 \\
ROW & 0.713 & - & 0.956 & 1.080 & 51.446 \\
Russia & 0.133 & - & - & 0.171 & 28.783 \\
Spain & 0.065 & - & - & 0.080 & 23.089 \\
United Kingdom & 0.114 & - & - & 0.118 & 4.149 \\
United States & 0.699 & - & - & 0.770 & 10.210 \\
WORLD & 2.093 & - & - & 2.887 & 37.943 \\
    \hline
    \end{tabular}
    \end{table}
\subsection*{Comparison of Results}
\begin{figure}[H]
    \centering
    \includegraphics[width=0.5\textwidth]{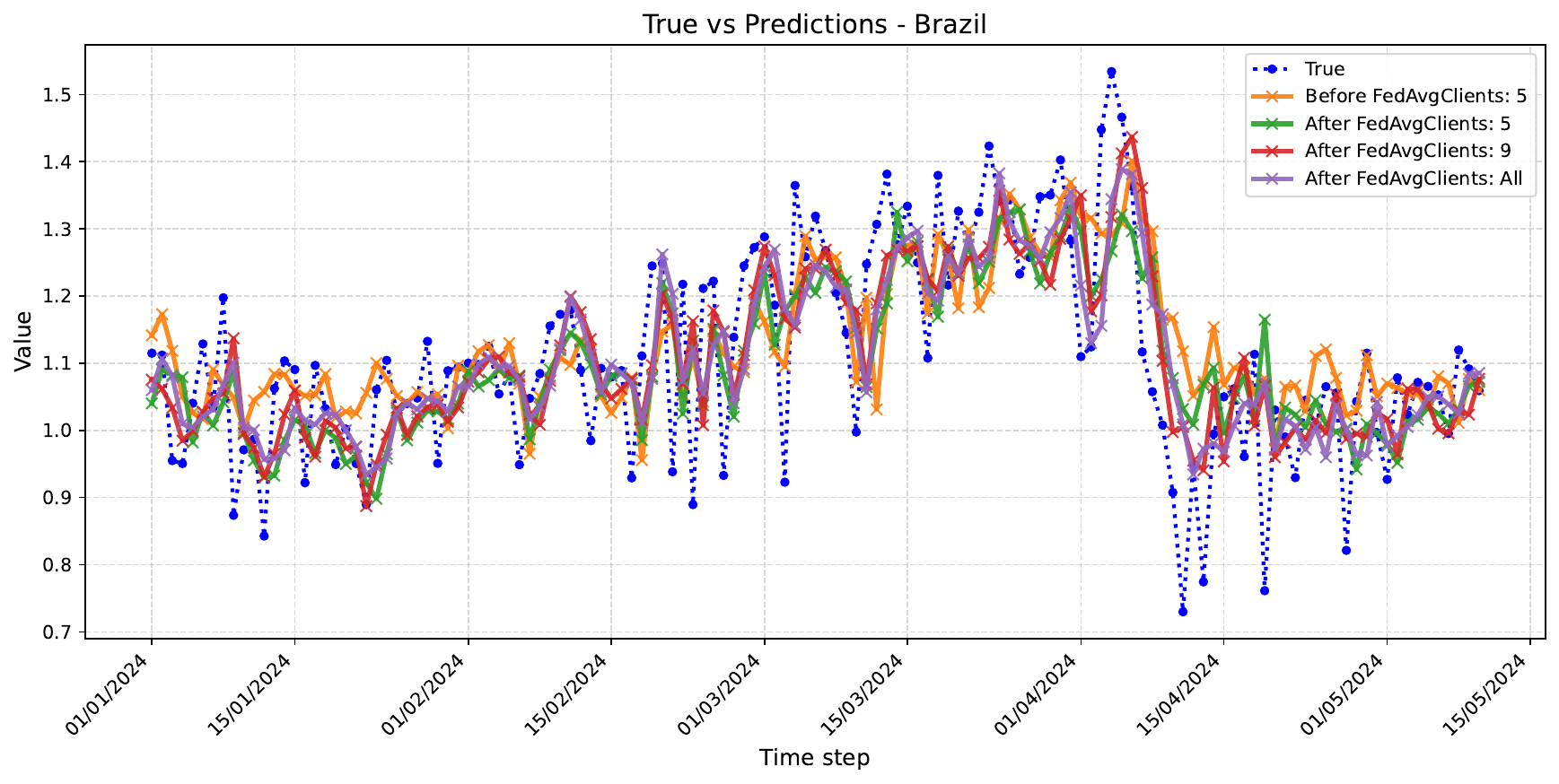}
    \caption{Brazil CO\textsubscript{2} emissions: 30\% data (Actual vs. FedAvg predictions).}
    \label{fig:brazil_plot_30}
\end{figure}

\begin{figure}[H]
    \centering
    \includegraphics[width=0.5\textwidth]{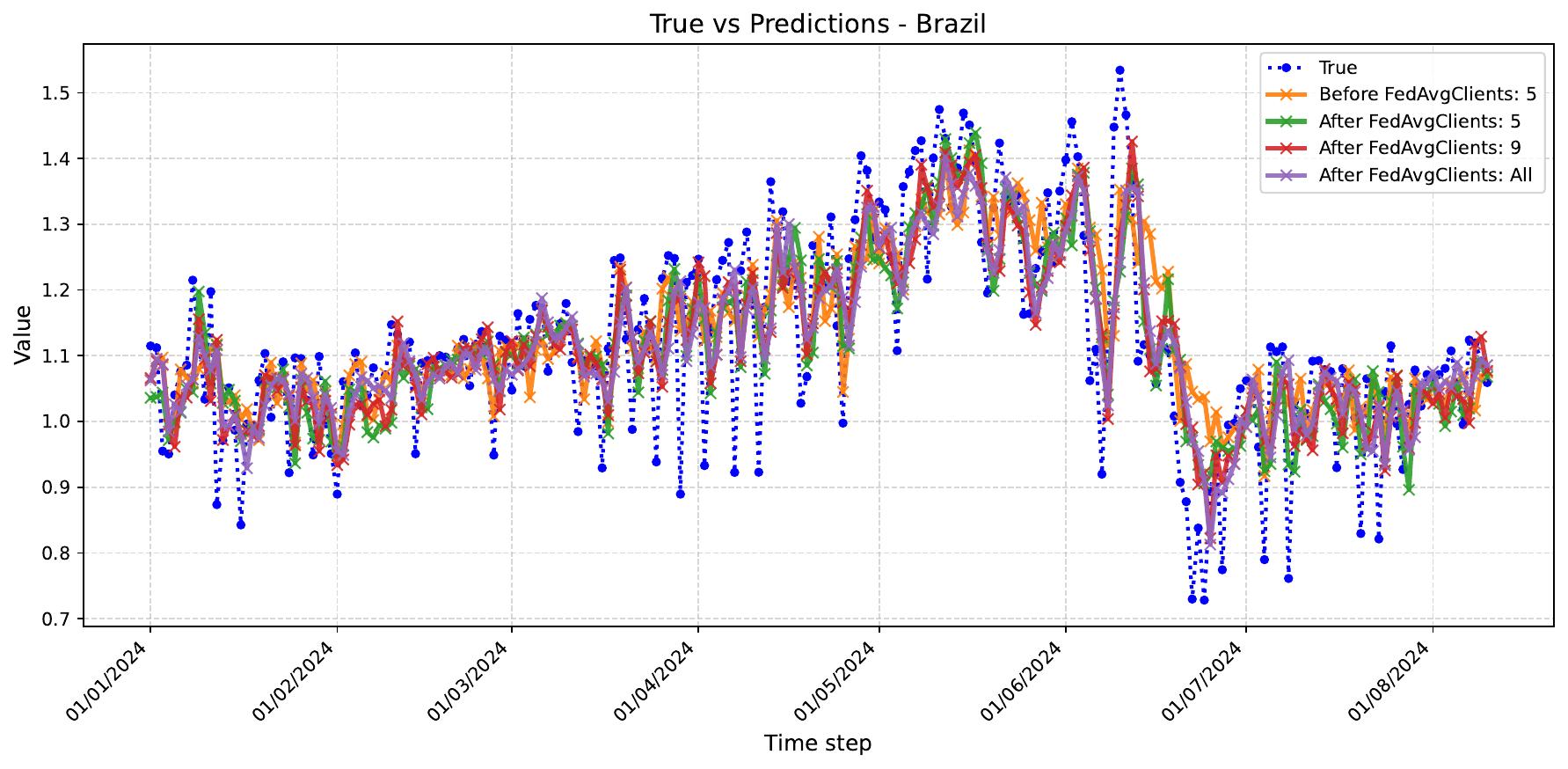}
    \caption{Brazil CO\textsubscript{2} emissions: 50\% data (Actual vs. FedAvg predictions).}
    \label{fig:brazil_plot_50}
\end{figure}

\begin{figure}[H]
    \centering
    \includegraphics[width=0.5\textwidth]{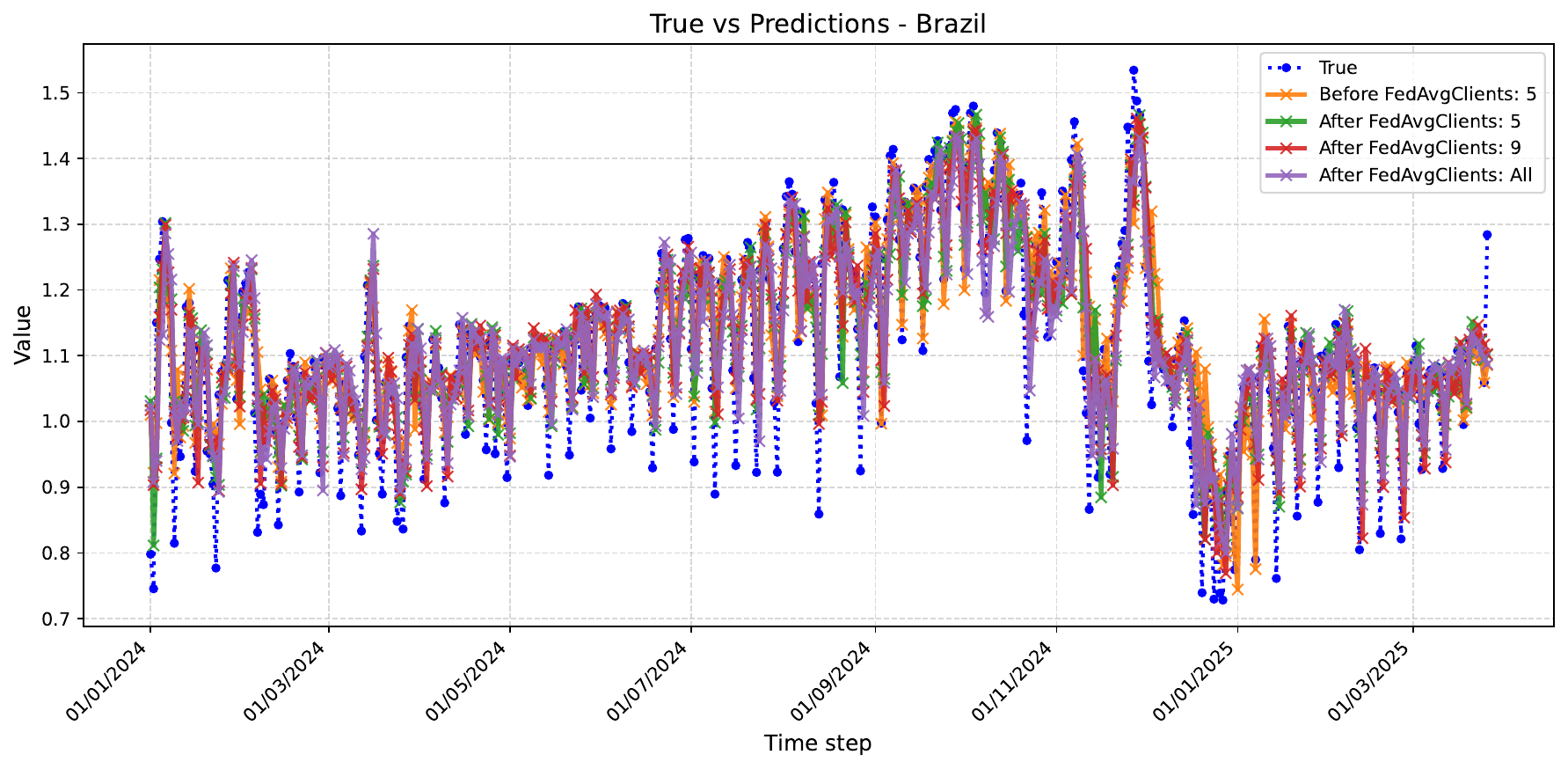}
    \caption{Brazil CO\textsubscript{2} emissions: 100\% data (Actual vs. FedAvg predictions).}
    \label{fig:brazil_plot_100}
\end{figure}

\begin{figure}[H]
    \centering
    \includegraphics[width=0.5\textwidth]{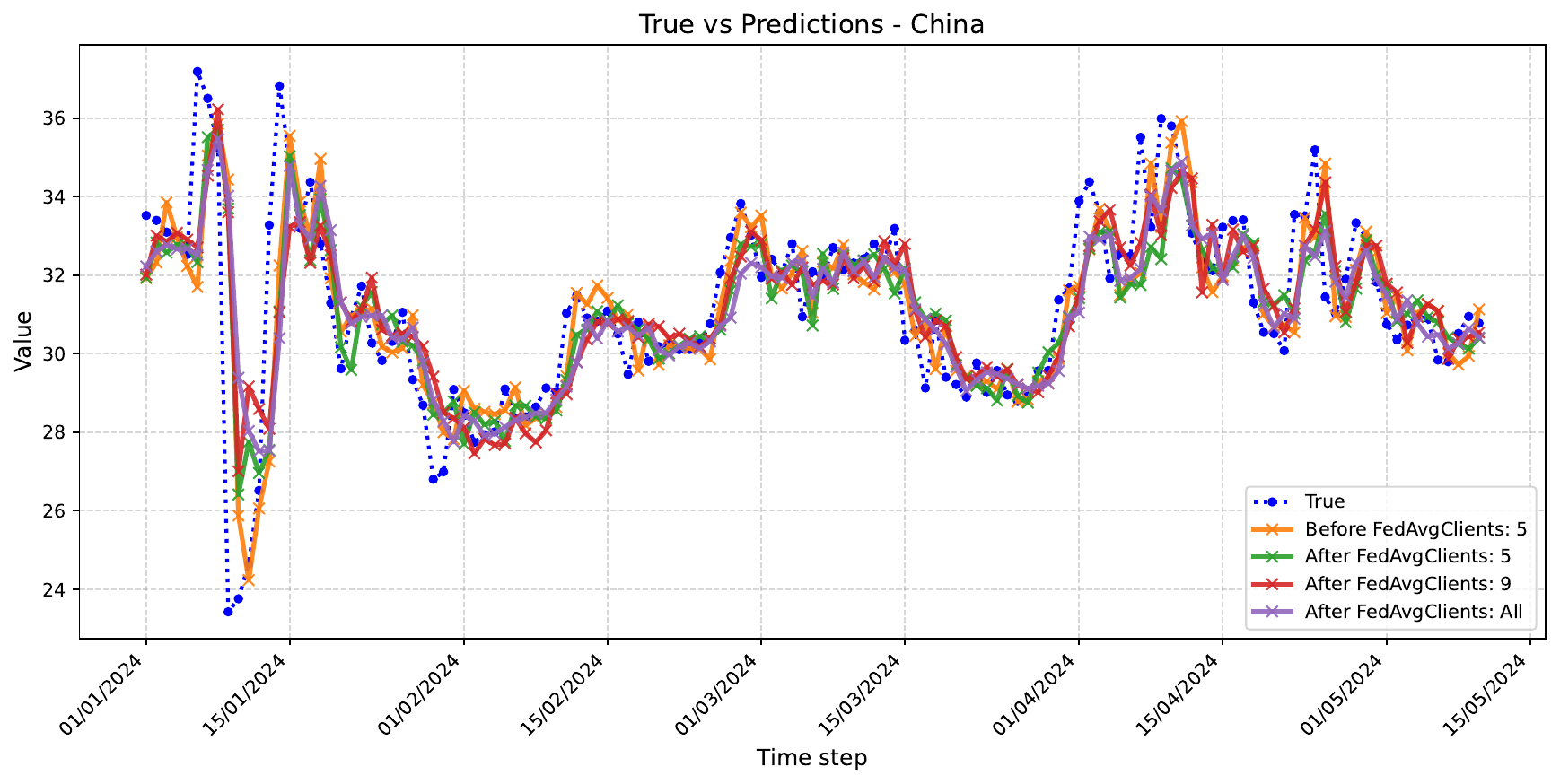}
    \caption{China CO\textsubscript{2} emissions: 30\% data (Actual vs. FedAvg predictions).}
    \label{fig:china_plot_30}
\end{figure}

\begin{figure}[H]
    \centering
    \includegraphics[width=0.5\textwidth]{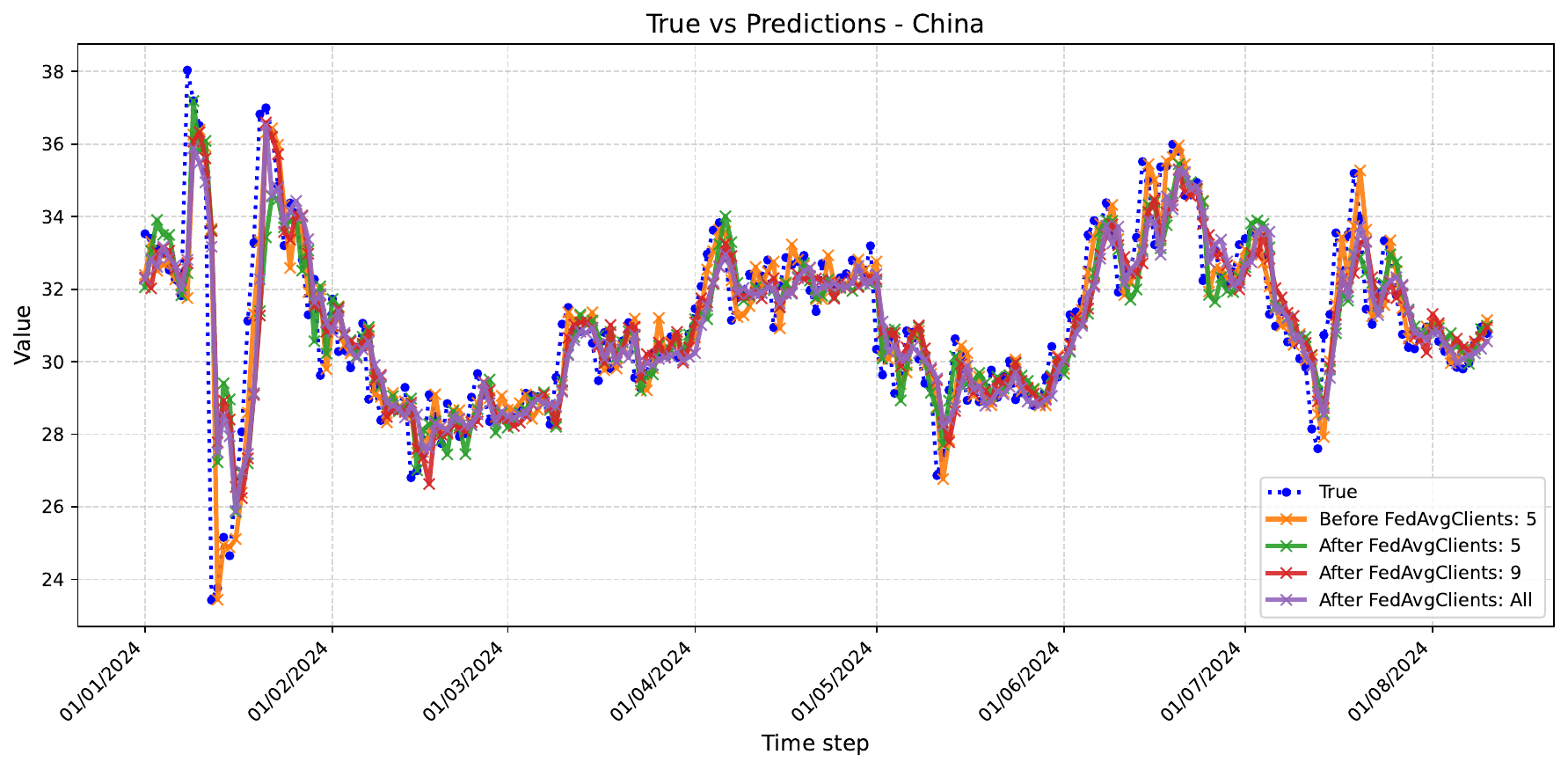}
    \caption{China CO\textsubscript{2} emissions: 50\% data (Actual vs. FedAvg predictions).}
    \label{fig:china_plot_50}
\end{figure}

\begin{figure}[H]
    \centering
    \includegraphics[width=0.5\textwidth]{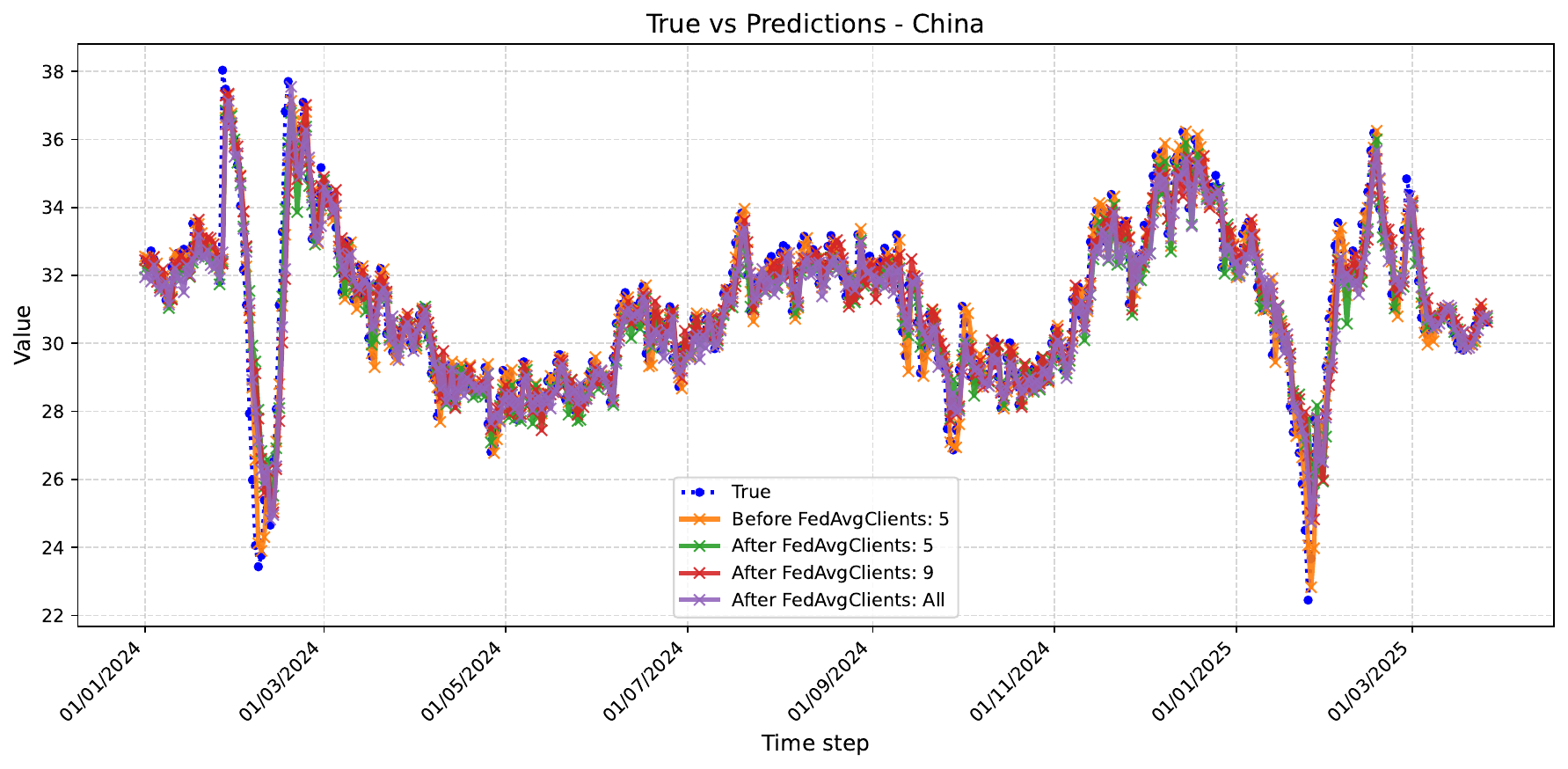}
    \caption{China CO\textsubscript{2} emissions: 100\% data (Actual vs. FedAvg predictions).}
    \label{fig:china_plot_100}
\end{figure}

\begin{figure}[H]
    \centering
    \includegraphics[width=0.5\textwidth]{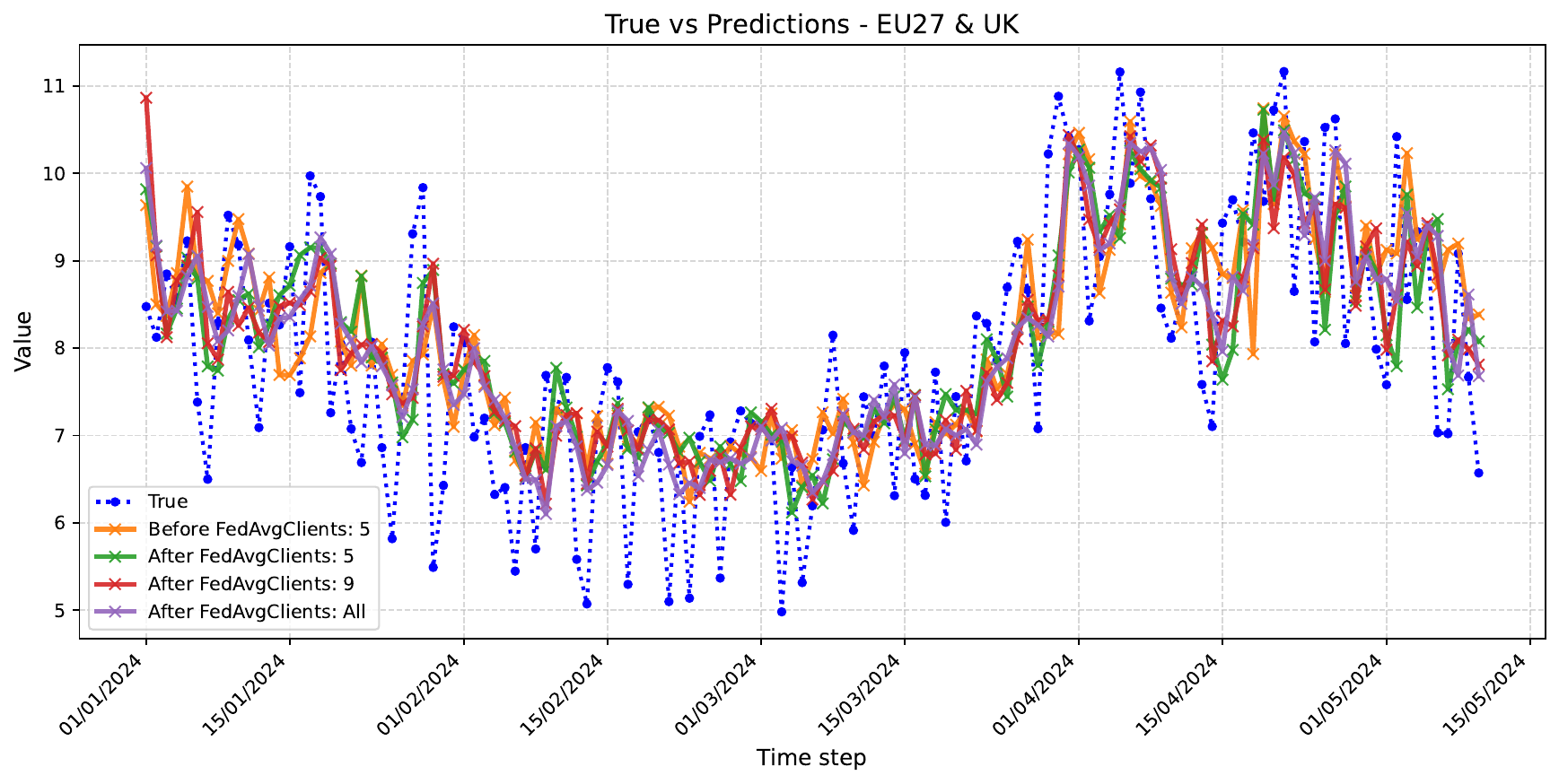}
    \caption{EU27 \& UK CO\textsubscript{2} emissions: 30\% data (Actual vs. FedAvg predictions).}
    \label{fig:eu27uk_plot_30}
\end{figure}

\begin{figure}[H]
    \centering
    \includegraphics[width=0.5\textwidth]{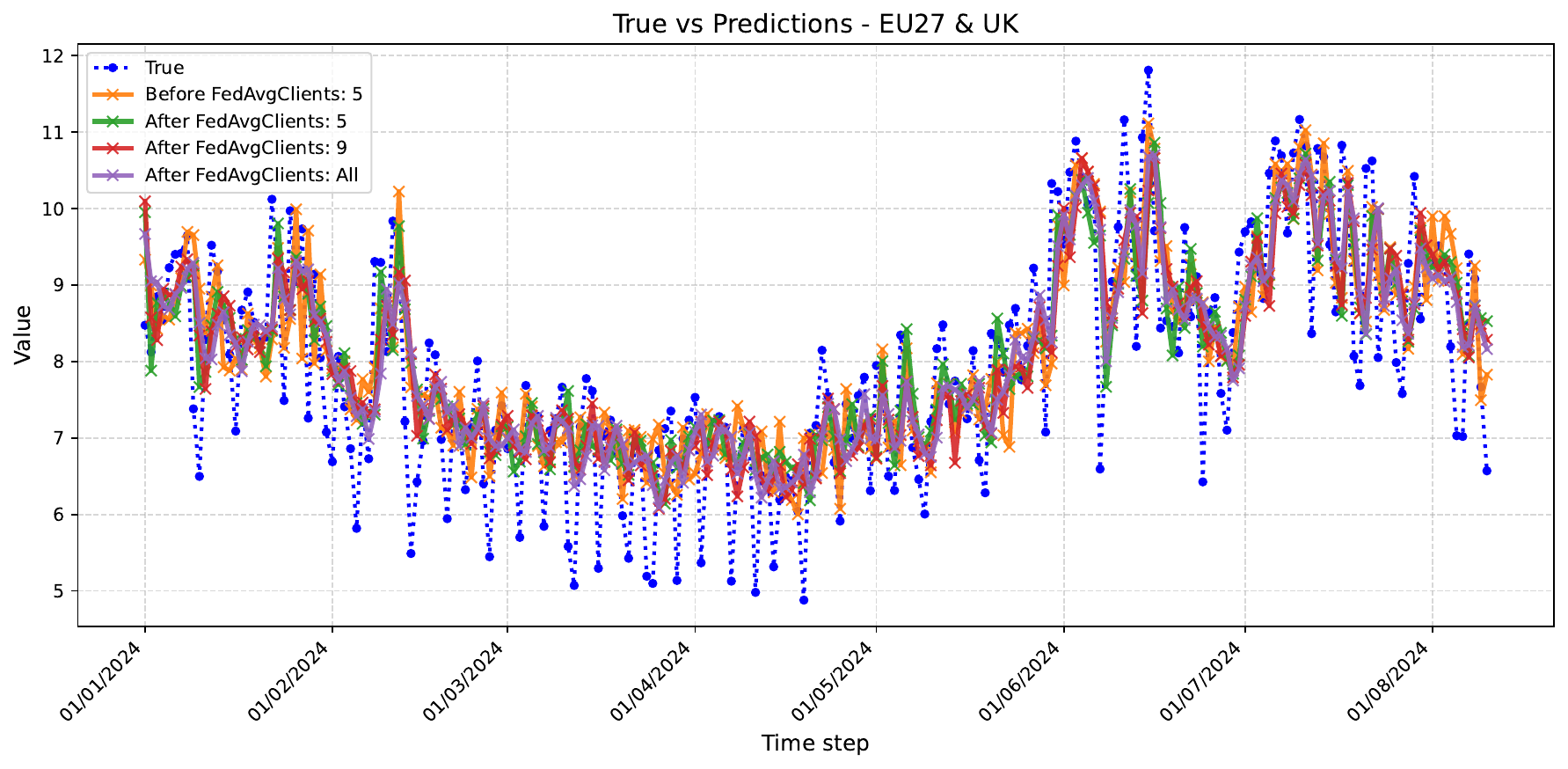}
    \caption{EU27 \& UK CO\textsubscript{2} emissions: 50\% data (Actual vs. FedAvg predictions).}
    \label{fig:eu27uk_plot_50}
\end{figure}

\begin{figure}[H]
    \centering
    \includegraphics[width=0.5\textwidth]{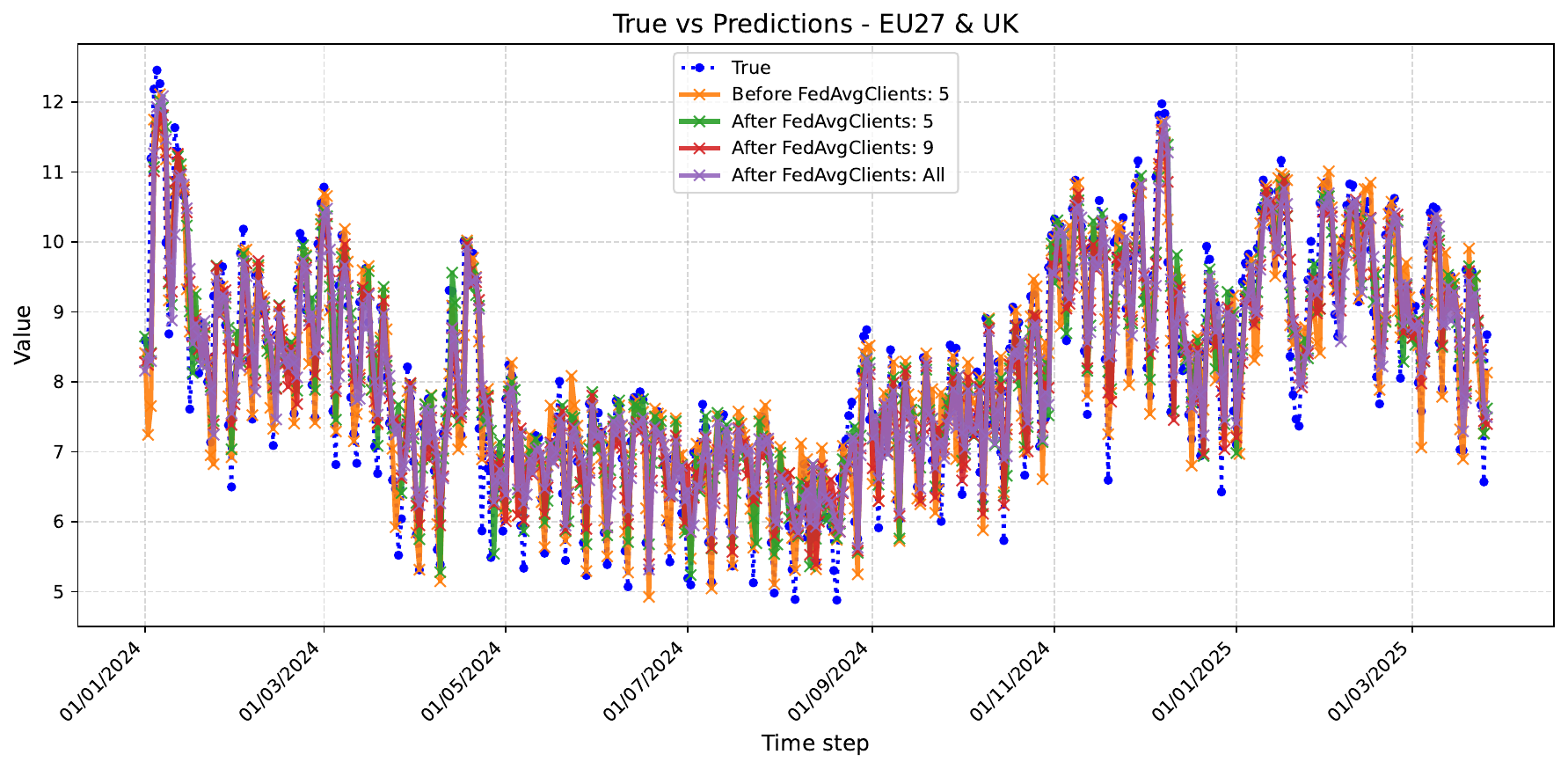}
    \caption{EU27 \& UK CO\textsubscript{2} emissions: 100\% data (Actual vs. FedAvg predictions).}
    \label{fig:eu27uk_plot_100}
\end{figure}

When comparing performance over the three data regimes (30\%, 50\%, and 100\%), a significant improvement trend emerges, particularly when moving from the smallest subset (30\%) to the mid-level dataset (50\%).  As expected, adding additional training data improves the model: the predicted series begins to closely track the real \texttt{total\_CO2} values, and the error values decline for most customers.  However, the scenario with 100\% data yields an intriguing result.  While numerical measures like $R^2$ occasionally imply only minor advances, the graphs (Fig. \ref{fig:brazil_plot_30}-\ref{fig:eu27uk_plot_100}) clearly illustrate that the forecasts after FedAvg nearly precisely track the actual CO\textsubscript{2} series.  A larger dataset affects the $R^2$ score due to a collection of small variations at each location. With a larger dataset, the $R^2$ score is influenced by the higher number of points. Small changes at each point stack up and can somewhat reduce the score, even if the overall fit seems fine.

\begin{figure}[H]
    \centering
    \includegraphics[width=0.5\textwidth]{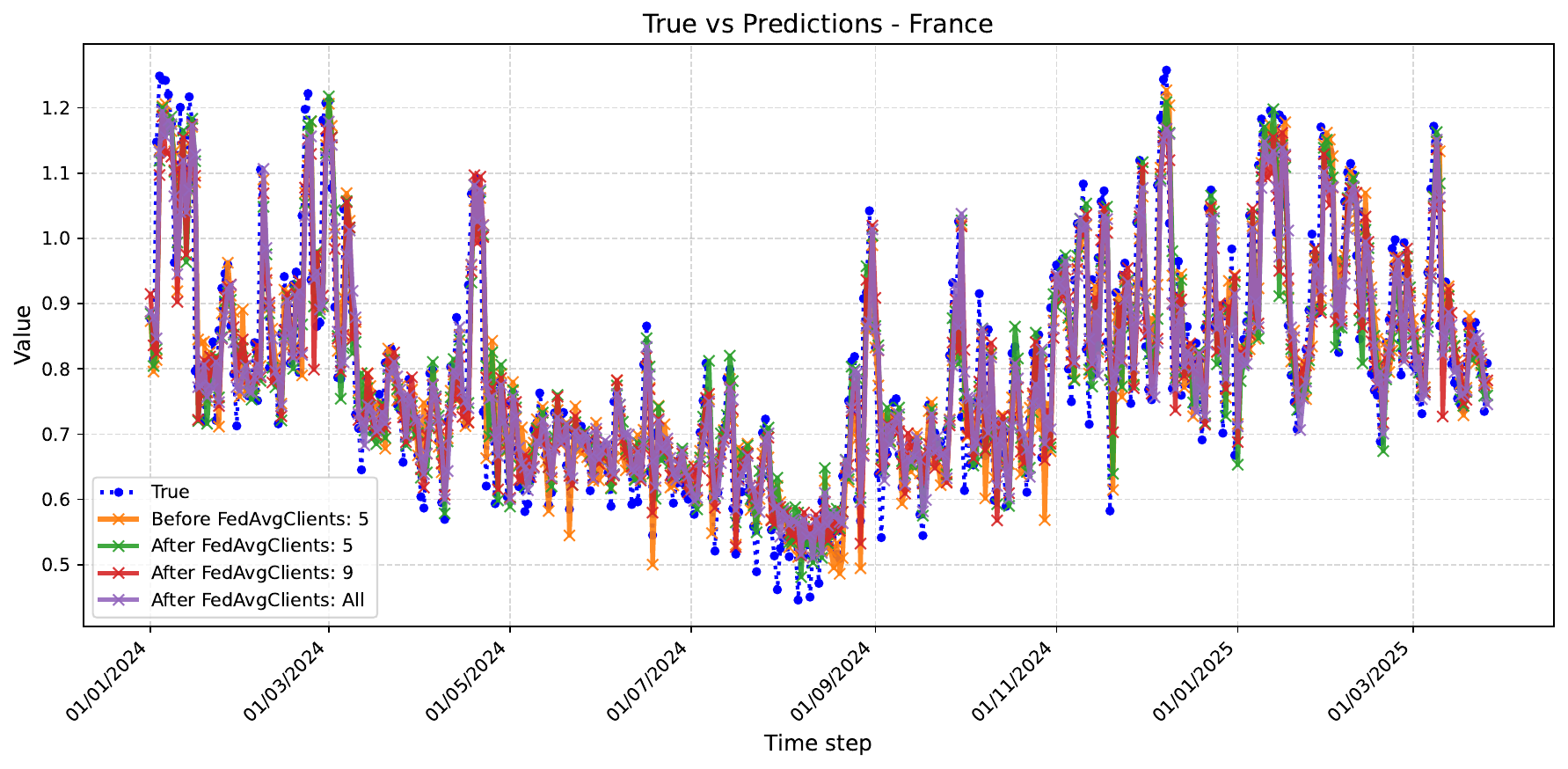}
    \caption{France CO\textsubscript{2} emissions: 100\% data (Actual vs. FedAvg predictions).}
    \label{fig:france_plot}
\end{figure}

\begin{figure}[H]
    \centering
    \includegraphics[width=0.5\textwidth]{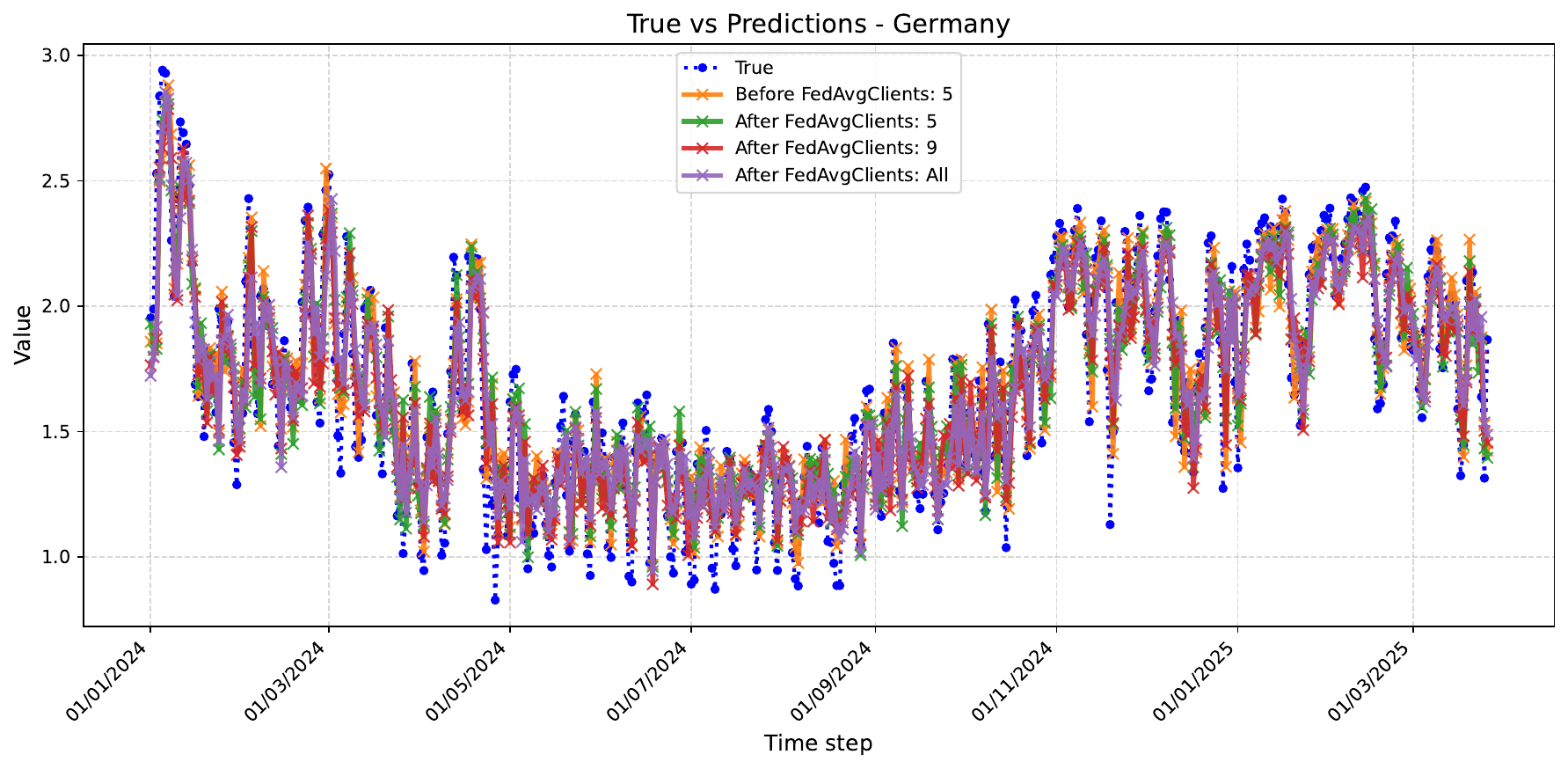}
    \caption{Germany CO\textsubscript{2} emissions: 100\% data (Actual vs. FedAvg predictions).}
    \label{fig:germany_plot}
\end{figure}

\begin{figure}[H]
    \centering
    \includegraphics[width=0.5\textwidth]{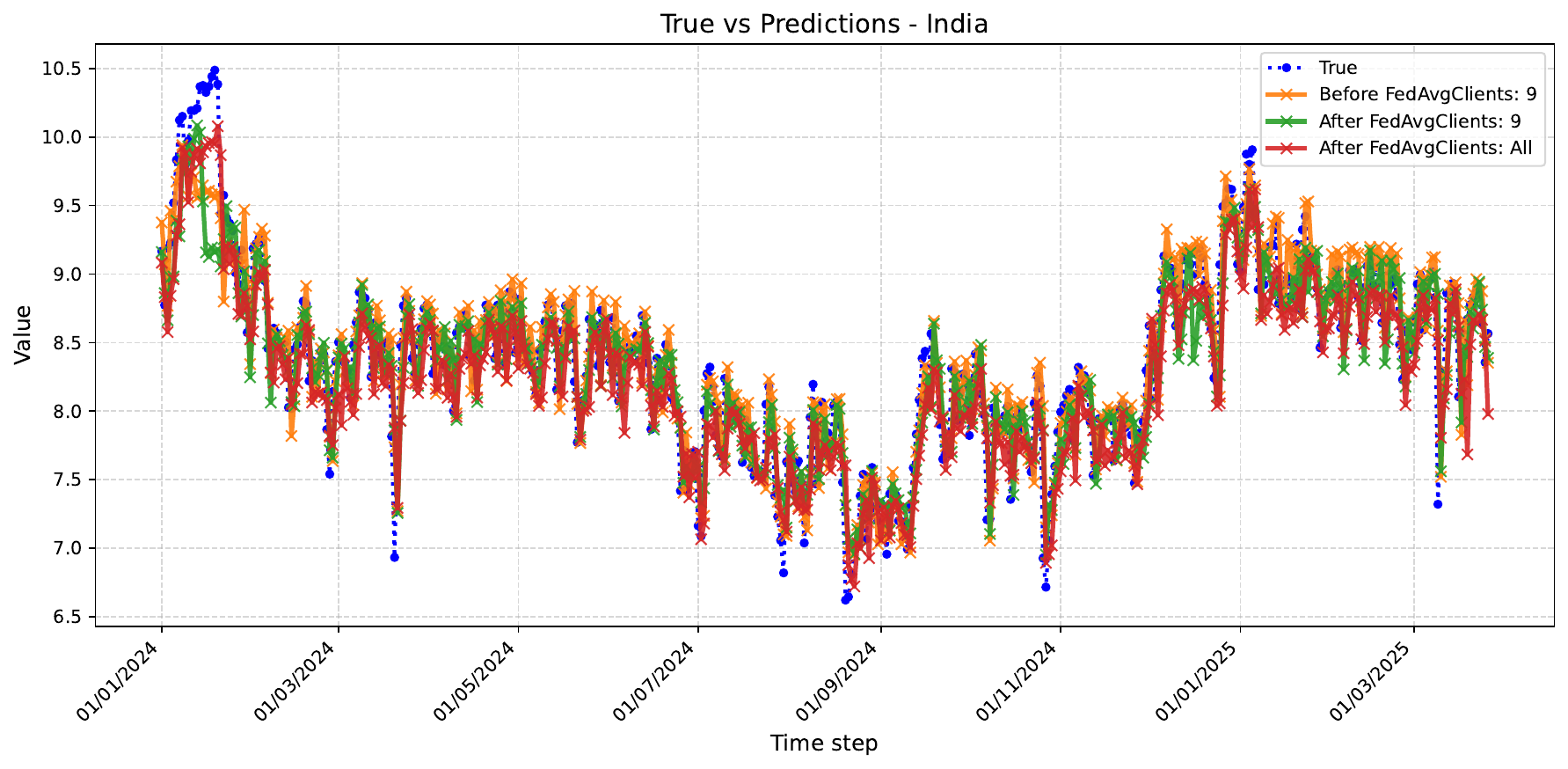}
    \caption{Actual vs. Predicted CO\textsubscript{2} emissions for India under different FedAvg scenarios.}
    \label{fig:india_plot}
\end{figure}

\begin{figure}[H]
    \centering
    \includegraphics[width=0.5\textwidth]{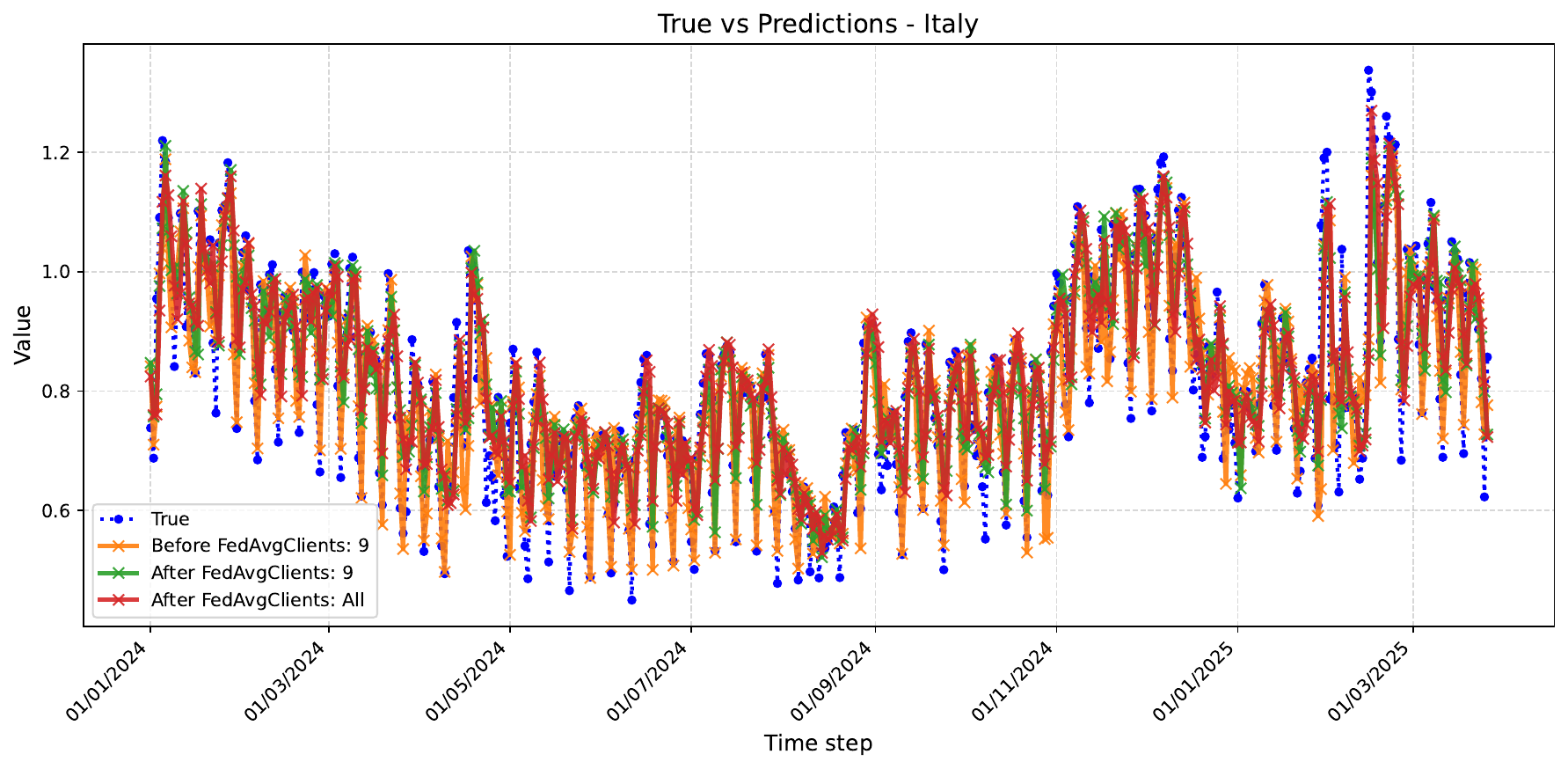}
    \caption{Actual vs. Predicted CO\textsubscript{2} emissions for Italy under different FedAvg scenarios.}
    \label{fig:italy_plot}
\end{figure}

\begin{figure}[H]
    \centering
    \includegraphics[width=0.5\textwidth]{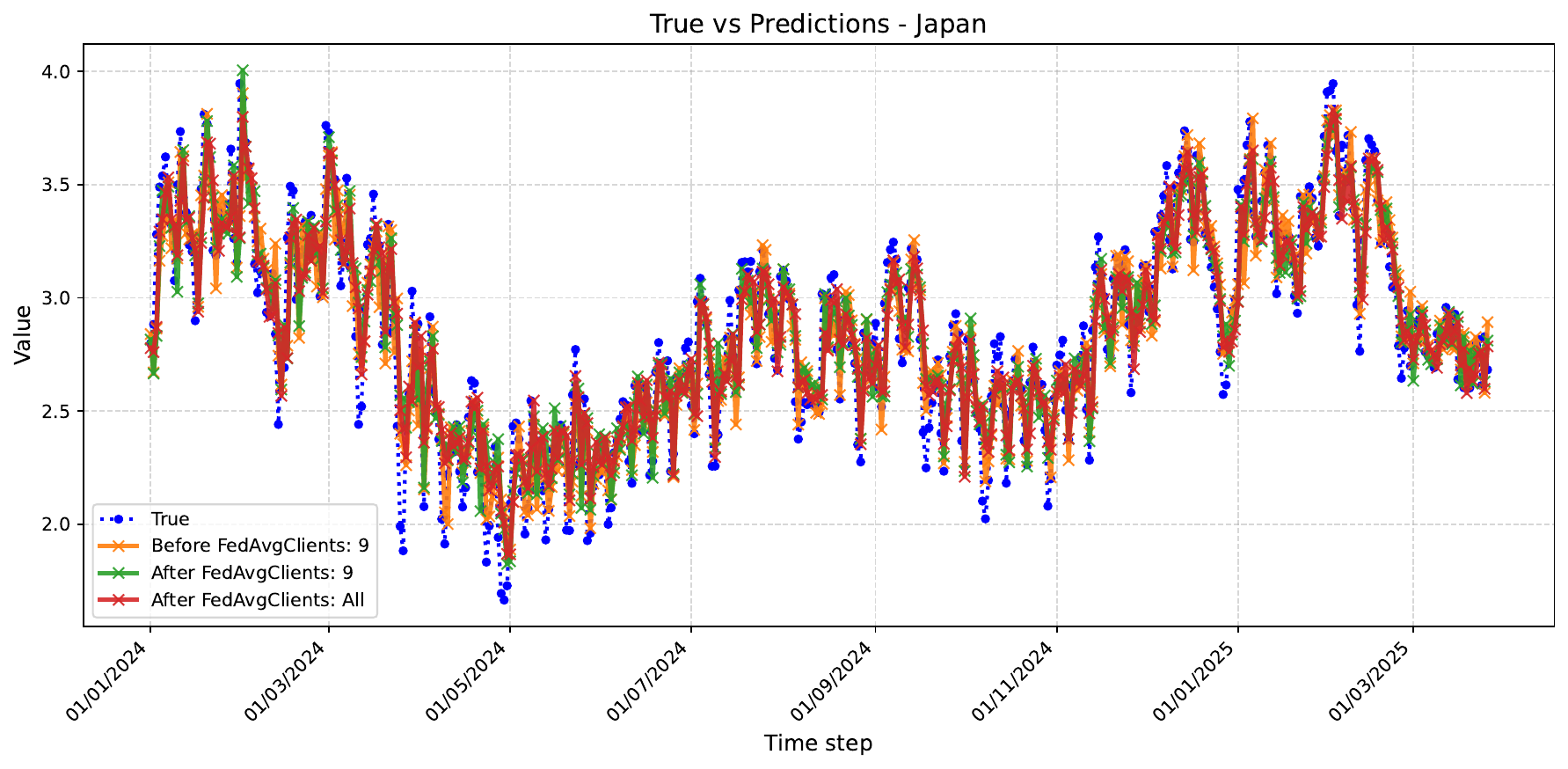}
    \caption{Actual vs. Predicted CO\textsubscript{2} emissions for Japan under different FedAvg scenarios.}
    \label{fig:japan_plot}
\end{figure}

\begin{figure}[H]
    \centering
    \includegraphics[width=0.5\textwidth]{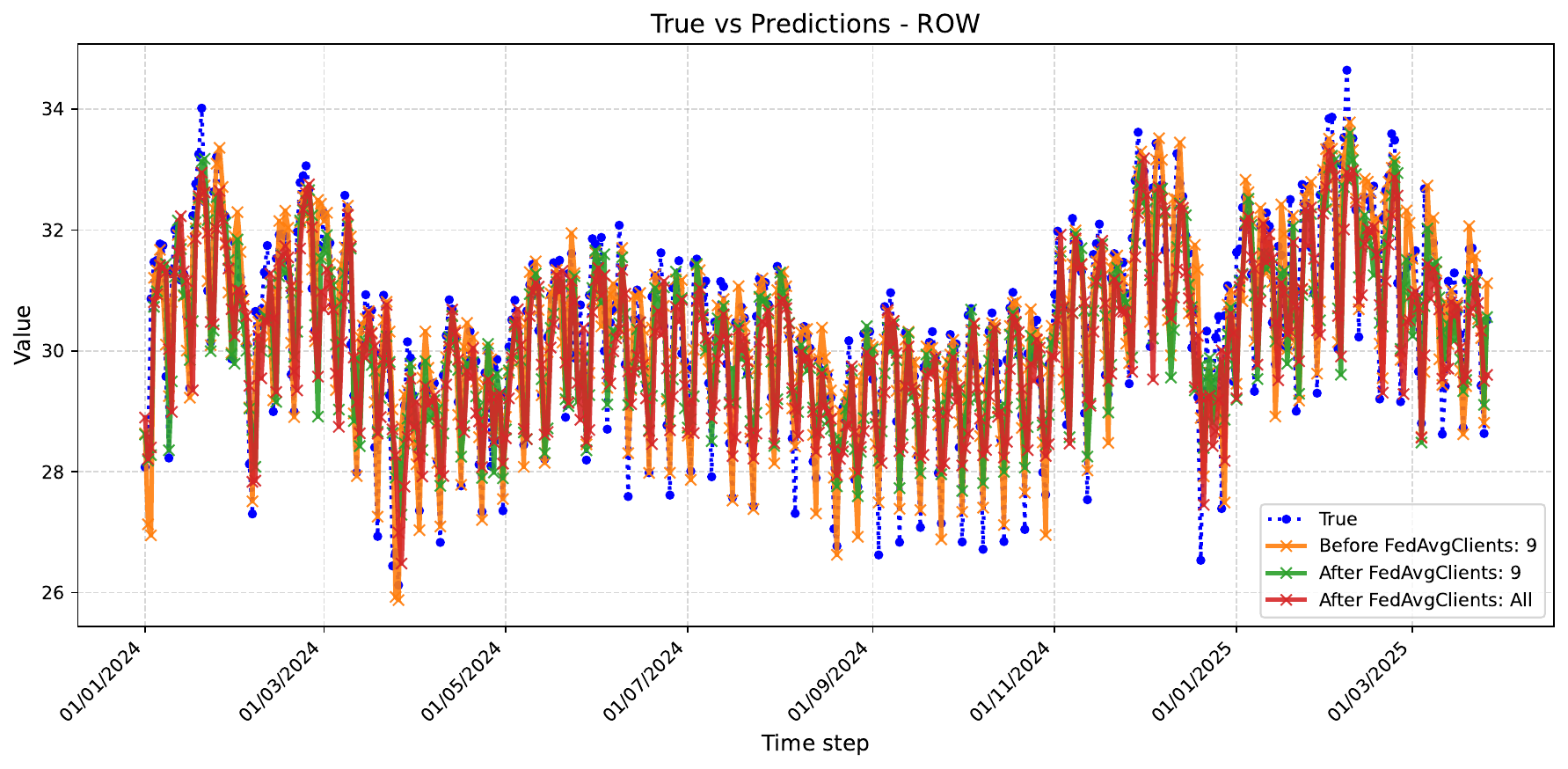}
    \caption{Actual vs. Predicted CO\textsubscript{2} emissions for Rest of the World (ROW) under different FedAvg scenarios.}
    \label{fig:row_plot}
\end{figure}

\begin{figure}[H]
    \centering
    \includegraphics[width=0.5\textwidth]{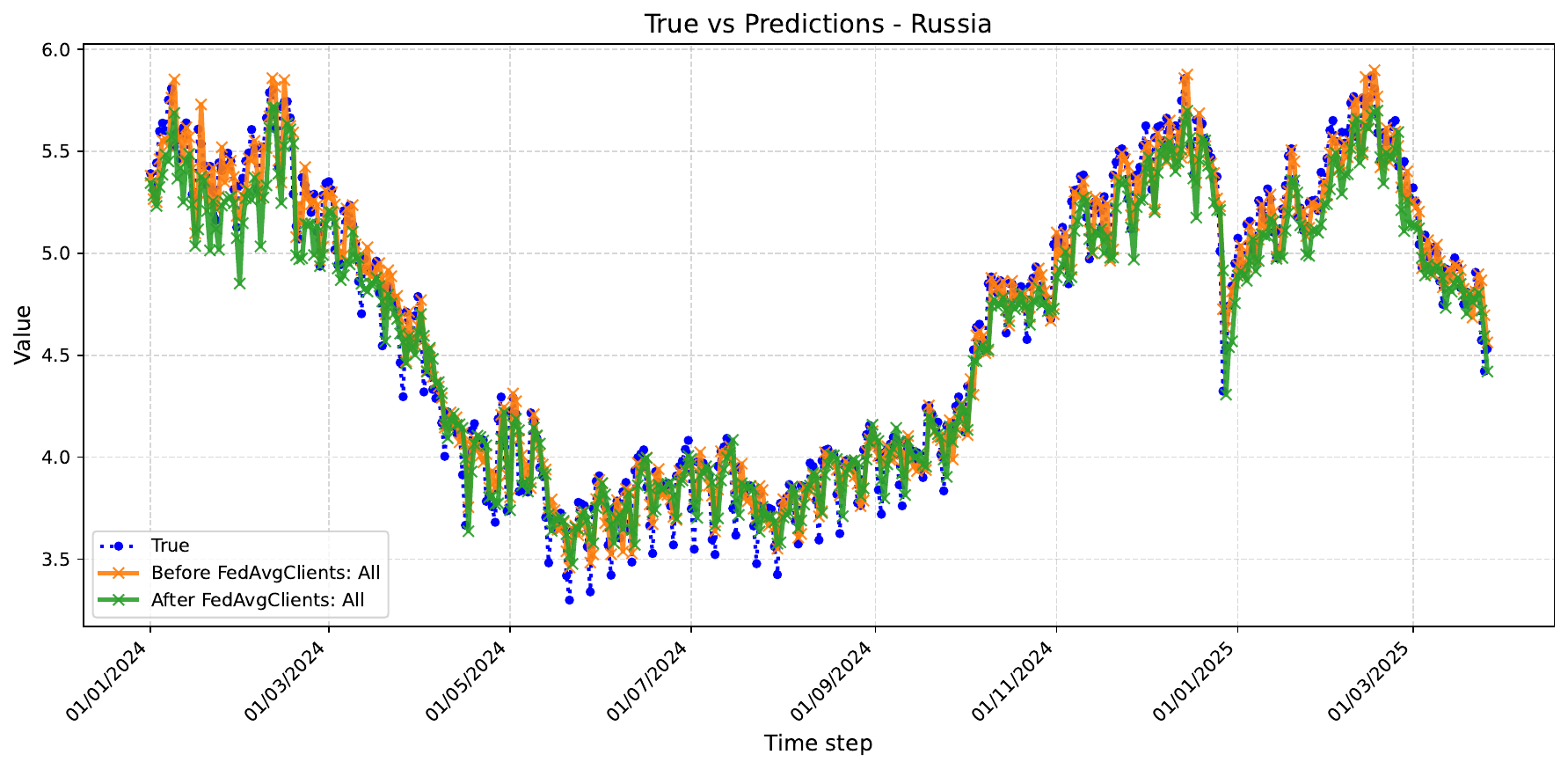}
    \caption{Actual vs. Predicted CO\textsubscript{2} emissions for Russia under different FedAvg scenarios.}
    \label{fig:russia_plot}
\end{figure}

\begin{figure}[H]
    \centering
    \includegraphics[width=0.5\textwidth]{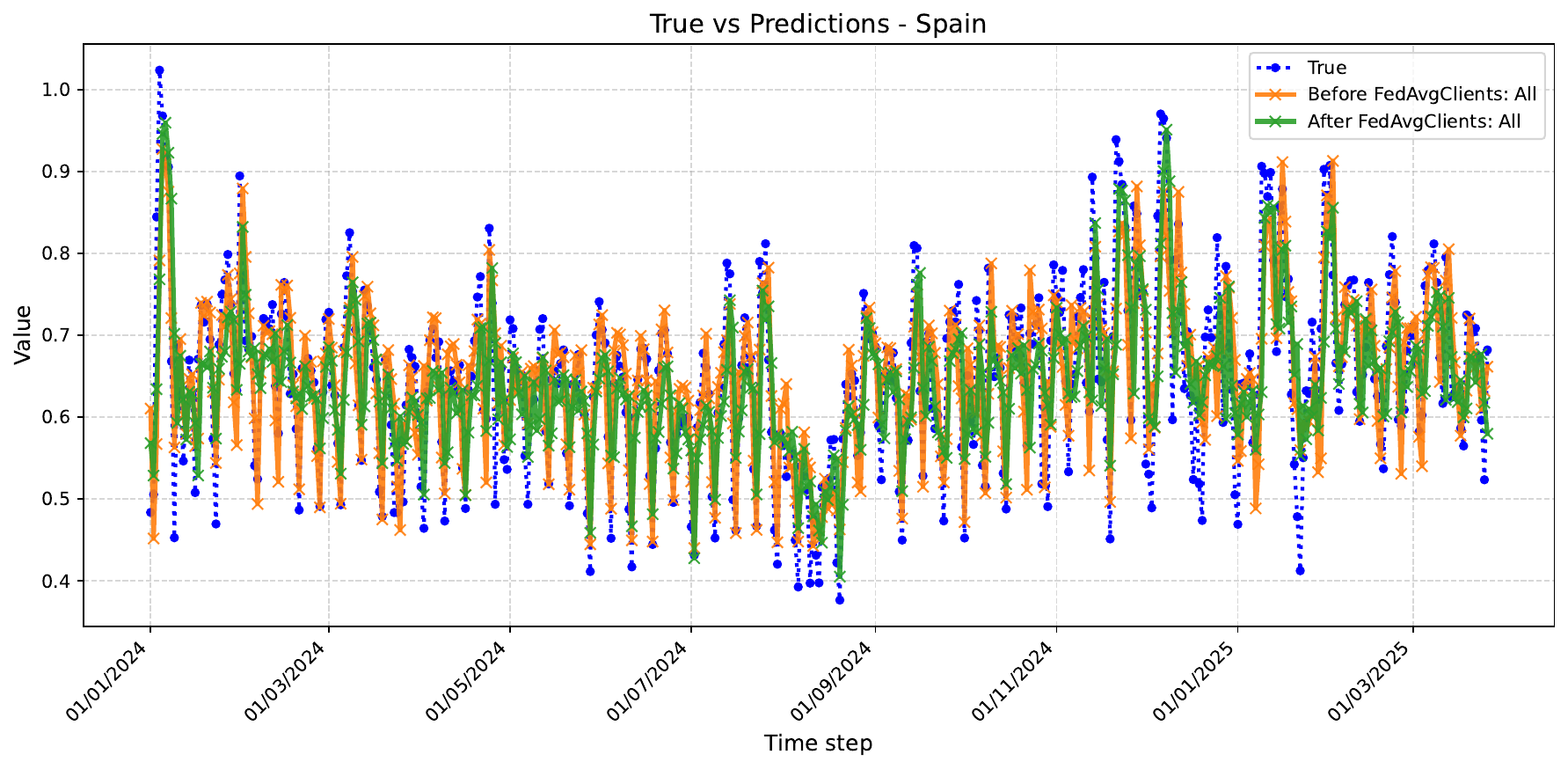}
    \caption{Actual vs. Predicted CO\textsubscript{2} emissions for Spain under different FedAvg scenarios.}
    \label{fig:spain_plot}
\end{figure}

\begin{figure}[H]
    \centering
    \includegraphics[width=0.5\textwidth]{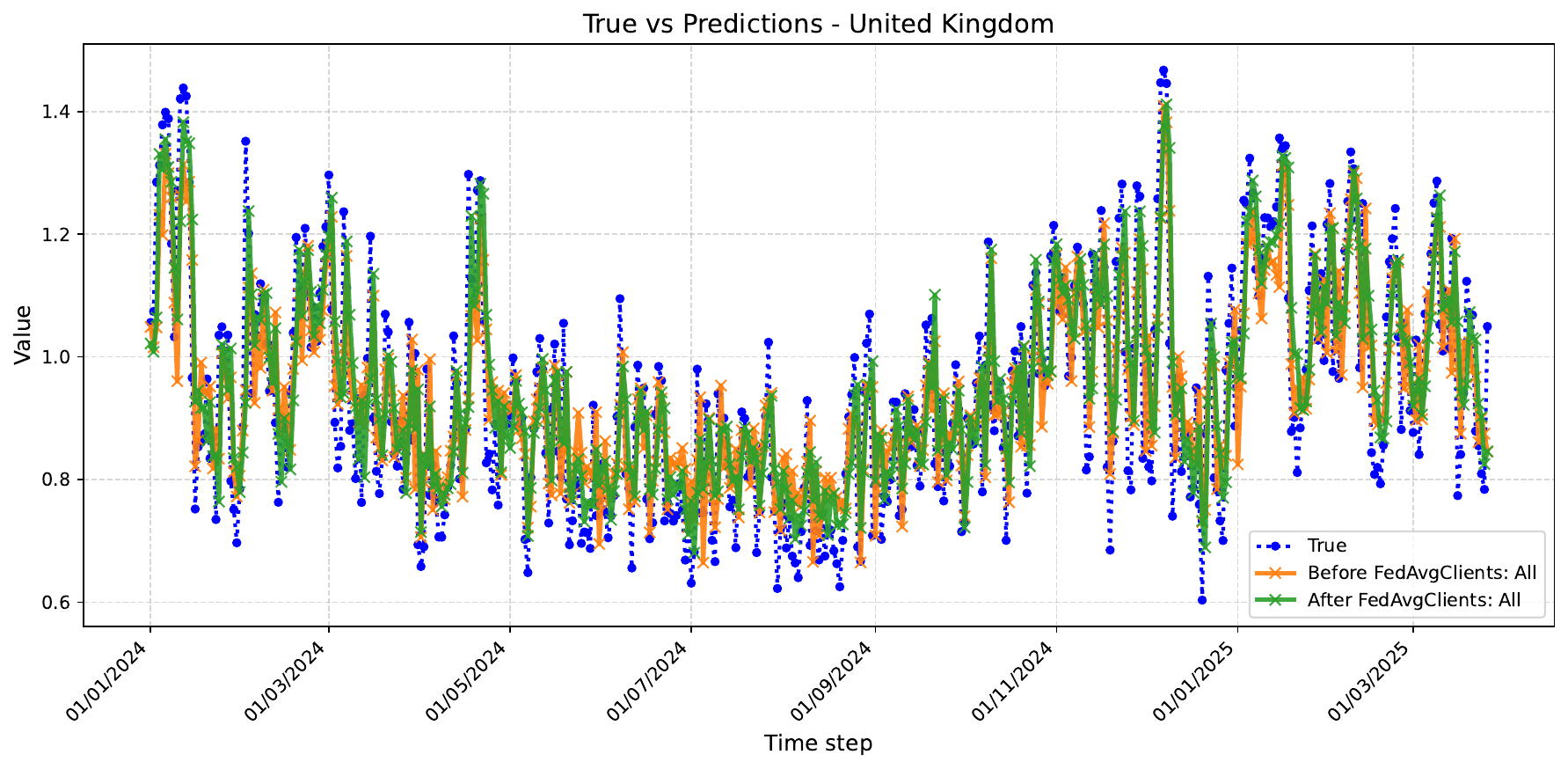}
    \caption{Actual vs. Predicted CO\textsubscript{2} emissions for the United Kingdom under different FedAvg scenarios.}
    \label{fig:uk_plot}
\end{figure}

\begin{figure}[H]
    \centering
    \includegraphics[width=0.5\textwidth]{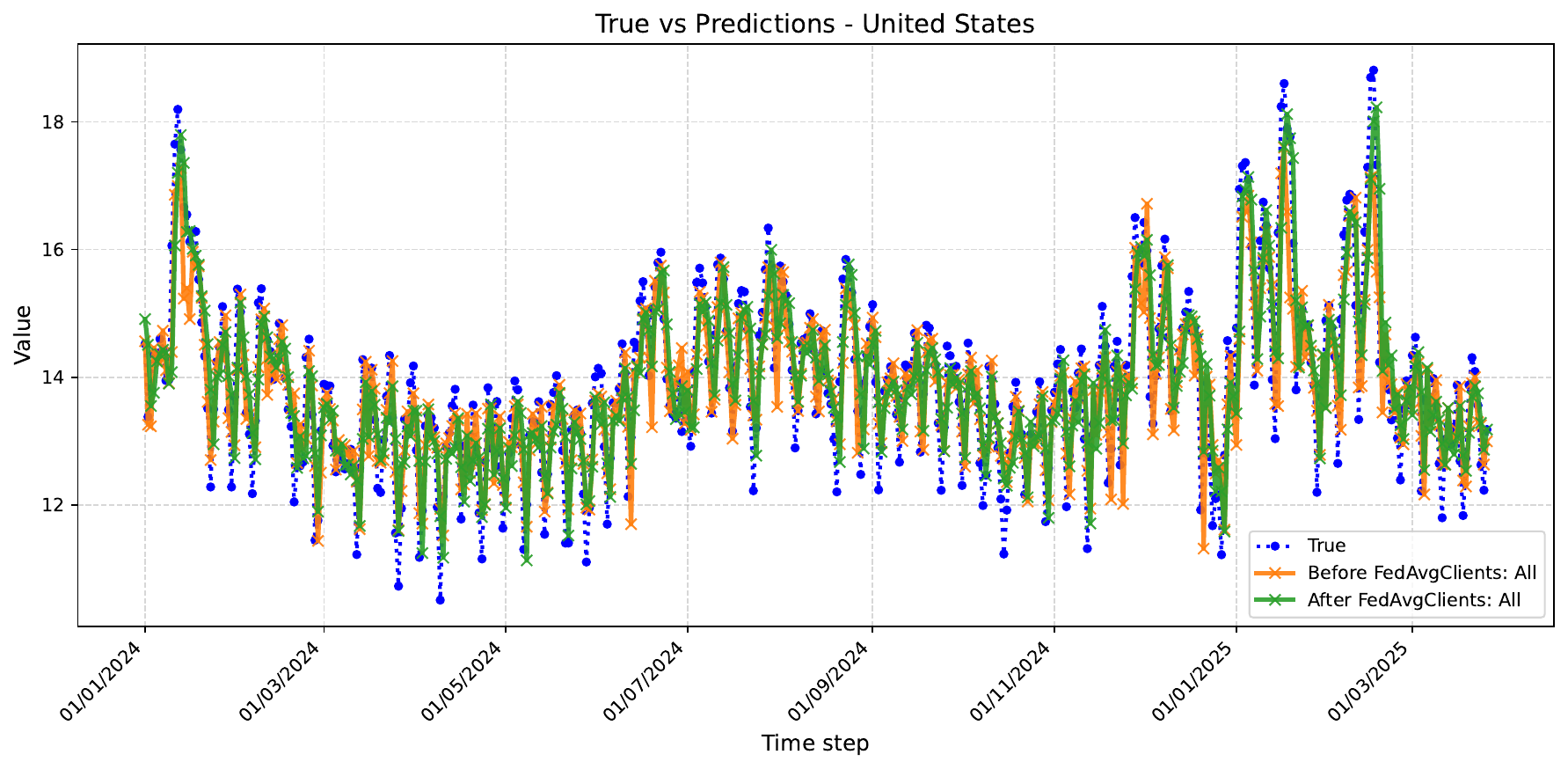}
    \caption{Actual vs. Predicted CO\textsubscript{2} emissions for the United States under different FedAvg scenarios.}
    \label{fig:us_plot}
\end{figure}

\begin{figure}[H]
    \centering
    \includegraphics[width=0.5\textwidth]{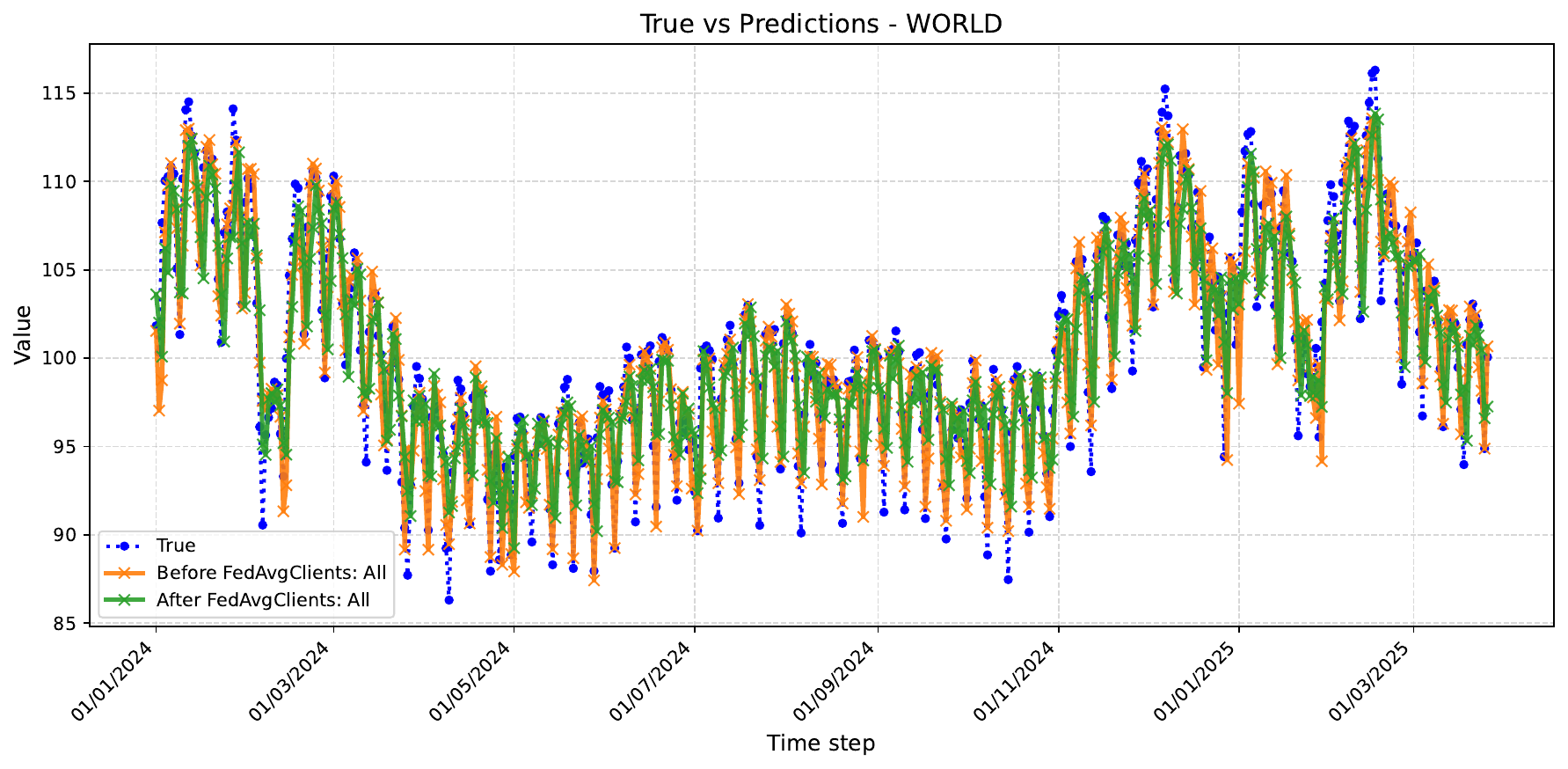}
    \caption{Actual vs. Predicted CO\textsubscript{2} emissions for the United States under different FedAvg scenarios.}
    \label{fig:world_plot}
\end{figure}

Using Brazil as an example (Fig. \ref{fig:brazil_plot_30}-\ref{fig:brazil_plot_100}), the 30\% plot shows the purple FedAvg line diverging from the blue-dotted actual series.  At 50\%, the line follows the data more closely, while there are still some gaps. At 100\%, the predicted and actual numbers are practically identical, indicating the best match.  The similar trend is found in China (Fig. \ref{fig:china_plot_30}-\ref{fig:china_plot_100}) and EU27 \& UK (Fig. \ref{fig:eu27uk_plot_30}-\ref{fig:eu27uk_plot_100}), with the 100\% case providing the most precise match between forecast and actual data.  For the other clients, the 100\% statistics (Fig. \ref{fig:france_plot}-\ref{fig:world_plot}) suggest that FedAvg forecasts can successfully capture the major fluctuations in CO\textsubscript{2} emissions

The results indicate that 30\% and 50\% datasets provide reasonable forecasts, but the 100\% dataset delivers the closest match with the actual CO\textsubscript{2} values. The below figures clearly show this progression more clearly. As the smaller data subsets leave visible gaps, while the full dataset brings the predicted and actual series almost on top of each other. Given the statement, it can be deduced that having more data is essential for stable and accurate forecasts in a federated setting.

% \onecolumn
% ------------------ Brazil ------------------

% ------------------ China ------------------

% ------------------ EU27 & UK ------------------

% ------------------ All other clients (100% only) ------------------

% \twocolumn
\subsection{Discussion}
\subsubsection{Challenges and Insights from Varying Client Participation}
The impact of client participation in federated aggregation is multifaceted. While adding clients to FedAvg is generally expected to enhance generalization by aggregating disparate patterns and reducing idiosyncratic noise, the observed results reveal nuanced dynamics. Our 30\% and 50\% illustrative tables show this tendency clearly. However, in the 100\% data scenario, client-specific responses become more pronounced: certain regions benefit from global aggregation, while others exhibit slightly lower $R^2$ and error metrics. Divergence can occur when: (i) the global model over-regularizes client-specific dynamics, (ii) clients exhibit non-overlapping seasonality or regime structure, making one model fits all unsatisfactory, or (iii) aggregation weights do not reflect sample size/variance. On contrary, there exists a great learning pattern among the visual graphs of the same data. So, it can be said that FL evaluation does not only rely on the statistical measures but also on the forecasting ability of the generalized model. 

However, in terms of statistical measures, below can be several factors that causes offset at full data:
\begin{enumerate}[label=(\roman*)]
\item \textbf{Client heterogeneity.} If sector wise emission differ sharply, a single global model may suppress high-signal features relevant to a subset of clients.
\item \textbf{Non-stationarity.} Structural breaks (policy changes, technology adoption, exogenous shocks) can cause ARIMA/GARCH parameters to differ across clients and periods, reducing the accuracy of computed averaging. However, excluding the peaks, the 100\% graphs show approximately accurate forecasting in almost all the clients.
\item \textbf{Aggregation weights.} Uniform weighting may not reflect client data volume or noise levels; adaptive strategies (e.g., FedMedian, FedProx, FedNova) or weighting by gradient variance could mitigate this.
\item \textbf{Target scaling.} Different client scales skew the global loss landscape.  Normalization per client or residualized series with z-scores can be useful in the given scenario.
\end{enumerate}

\subsubsection{Role of ARIMA/GARCH Features}
Including ARIMA and GARCH-derived features adds local autoregressive structure and conditional variance indicators to the FL pipeline.  These features are particularly useful in low-data regimes (30\%, 50\%), when direct comprehensive training may be variance-limited.  At 100\%, the incremental benefit may be muted for some clients, but remains useful for volatility-prone series.

\subsubsection{Practical Guidance}
The adopted approach can depict following benefits in the given setup of experiments:
\begin{itemize}
\item Used client-wise normalization and sector-share covariates to align targets before aggregation.
\item Considered personalization layers or fine-tuning to recapture client-specific structure after global averaging.
\item Validate with rolling-origin splits to ensure robustness to temporal drift, and report both point metrics (MSE, MAE, RMSE, MAPE) and interval metrics where possible.
\end{itemize}

\subsubsection{Applicability Beyond \texttt{total\_CO$_2$}}
Although demonstrated on \texttt{total\_CO$_2$}, the framework is sector-agnostic and can be applied to any sectoral time series after the same row-to-column transformation.

\section{Conclusion}
This study effectively addressed the significant issue of predicting CO\textsubscript{2} emissions without consolidating sensitive, scattered data by presenting an innovative federated learning approach.  The proposed hybrid model integrates ARIMA and GARCH for statistical feature extraction with LSTM-Attention and XGBoost for nonlinear pattern recognition, efficiently capturing intricate temporal trends and volatility inside a privacy-preserving FedAvg protocol.  Our extensive experimental investigation showed the framework's resilience and scalability, exhibiting consistent performance enhancements across diverse data volumes and client setups.  The findings confirm that collaborative learning from decentralized data sources can produce accurate and dependable forecasts, essential for guiding sustainable policy decisions.  In future endeavors, we intend to investigate adaptive client weighting methodologies to more effectively address data heterogeneity, integrate external economic and policy variables, and expand the framework to encompass real-time forecasting situations.  Moreover, examining the environmental consequences of the FL communication overhead is crucial to guarantee the comprehensive sustainability of the solution.

\bibliographystyle{IEEEtran}
\bibliography{references}

@article{xiong2024,
  author={Xiong, Ao and Zhou, Han and Song, Yu and Wang, Dong and Wei, Xu and Li, Da and Gao, Bo},
  year={2024},
  title={A Multi-Task Based Clustering Personalized Federated Learning Method},
  journal={Big Data Mining and Analytics},
  volume={7},
  number={4},
  pages={1017--1030},
  doi={10.26599/BDMA.2024.9020001}
}

@article{qiu2023first,
  title   = {A First Look into the Carbon Footprint of Federated Learning},
  author  = {Qiu, Xinchi and Parcollet, Titouan and Fernandez-Marques, Javier and Gusmao, Pedro P. B. and Gao, Yan and Beutel, Daniel J. and Topal, Taner and Mathur, Akhil and Lane, Nicholas D.},
  journal = {Journal of Machine Learning Research},
  volume  = {24},
  pages   = {1--23},
  year    = {2023},
  url     = {http://jmlr.org/papers/v24/21-0445.html}
}

@inproceedings{mustafa2024forecasting,
  title     = {Forecasting Carbon Dioxide Emissions and Energy Sources in Bangladesh Using Statistical and Machine Learning Models},
  author    = {Mustafa, Md Asif and Marma, Mongsathowai and Haq, Md. Mahfujul and Hossain, Md. Jakir and Barua, Nejum},
  booktitle = {Proceedings of the 7th Bangladesh Conference on Industrial Engineering and Operations Management},
  publisher = {IEOM Society International},
  year      = {2024},
  pages     = {1330--1338},
  doi       = {10.46254/BA07.20240226}
}

@article{cui2023,
  author={Cui, Tianxu and Shi, Ying and Lv, Bo and Ding, Rijia and Li, Xianqiang},
  year={2023},
  title={Federated learning with SARIMA-based clustering for carbon emission prediction},
  journal={Journal of Cleaner Production},
  volume={426},
  pages={139069},
  doi={10.1016/j.jclepro.2023.139069}
}

@techreport{ipcc2024,
  author={IPCC},
  year={2024},
  title={Climate Change 2024: Synthesis Report. Contribution of Working Groups I, II and III},
  institution= {IPCC, Geneva, Switzerland},
  publisher={IPCC, Geneva, Switzerland}
}

@article{smith2017,
  author={Smith, Virginia and Chiang, Christopher-Kai and Sanjabi, Maziar and Talwalkar, Ameet},
  year={2017},
  title={Federated multi-task learning},
  journal={NeurIPS},
  volume={30}
}

@article{chen2020,
  author={Chen, Yang and Qin, Xiang and Wang, Jiawei and Yu, Bo and Gao, Wei},
  year={2020},
  title={FedHealth: Federated transfer learning for wearable healthcare},
  journal={IEEE Intelligent Systems},
  volume={35},
  number={4},
  pages={83--93}
}

@inproceedings{fallah2020,
  author={Fallah, Alireza and Mokhtari, Aryan and Ozdaglar, Asuman},
  year={2020},
  title={Personalized federated learning: A meta-learning approach},
  booktitle={NeurIPS},
  volume={33},
  pages={16513--16524}
}

@inproceedings{khodak2019,
  author={Khodak, Mikhail and Balcan, Maria-Florina and Talwalkar, Ameet},
  year={2019},
  title={Adaptive gradient-based meta-learning methods},
  booktitle={NeurIPS},
  volume={32}
}

@article{yu2021,
  author={Yu, Tao and Li, Tian and Sun, Shiqiang and Xu, Yao and Tao, Dacheng and Yang, Qiang},
  year={2021},
  title={FedKD: Communication efficient federated learning via knowledge distillation},
  journal={NeurIPS},
  volume={34}
}

@article{zhu2021,
  author={Zhu, Zihan and Hong, Zhang and Xu, Jiale and Wang, Qiang and Yang, Qiang},
  year={2021},
  title={MetaFed: Federated learning with meta-knowledge},
  journal={KDD},
  pages={237--246}
}

@article{long2022,
  title={Multi-center federated learning: clients clustering for better personalization},
  author={Long, Guodong and Xie, Ming and Shen, Tao and Zhou, Tianyi and Wang, Xianzhi and Jiang, Jing},
  journal={World Wide Web},
  volume={26},
  number={1},
  pages={481--500},
  year={2023},
  publisher={Springer}
}

@article{yoo2021,
  author={Yoo, Jinyoung and Kim, Jaehoon and Kim, Sang-Wook},
  year={2021},
  title={Personalized federated clustering for depression detection},
  journal={IEEE JBHI},
  volume={25},
  number={12},
  pages={4541--4552}
}

@article{appiah2018,
  author={Appiah, Michael},
  year={2018},
  title={Causality between CO2 emissions, energy consumption, economic growth and industrialization: Evidence from Sub-Saharan Africa},
  journal={Energy},
  volume={135},
  pages={1049--1069}
}

@article{wang2017,
  author={Wang, Shuang and Lin, Boqiang},
  year={2017},
  title={CO2 emissions projection in China based on VAR-STIRPAT model},
  journal={Energy Policy},
  volume={102},
  pages={601--612}
}

@article{yang2020,
  author={Yang, Yuting and O’Connell, John F},
  year={2020},
  title={Forecasting CO2 emissions of air transport industry in Shanghai: A multivariate ARIMA model},
  journal={J. Air Transport Management},
  volume={87},
  pages={101856}
}

@article{niu2020,
  author={Niu, Dejian and Wang, Qiang and Wu, Wei and Zhao, Xian},
  year={2020},
  title={Forecasting carbon emissions using an improved fireworks algorithm and GRNN},
  journal={Journal of Cleaner Production},
  volume={276},
  pages={124120}
}

@article{gao2021,
  author={Gao, Min and Sun, Yu and Li, Hui},
  year={2021},
  title={A fractional grey Riccati model for CO2 emissions estimation},
  journal={Journal of Cleaner Production},
  volume={279},
  pages={123456}
}

@article{huang2019,
  author={Huang, Yijie and Li, Xia and Wang, Qiang},
  year={2019},
  title={Carbon emission prediction using LSTM and grey correlation analysis},
  journal={Applied Energy},
  volume={240},
  pages={619--632}
}

@article{savazzi2023energy,
  title   = {An energy and carbon footprint analysis of distributed and federated learning},
  author  = {Savazzi, Stefano and Rampa, Vittorio and Kianoush, Sanaz and Bennis, Mehdi},
  journal = {IEEE Transactions on Green Communications and Networking},
  volume  = {7},
  number  = {1},
  pages   = {248--264},
  year    = {2023}
}

@book{islam2024data,
  title={Data-Driven Approaches for Achieving Carbon Neutrality: Predictive Models for Reducing CO2 Emissions and Enhancing Industrial Sustainability},
  author={Islam, Farzana},
  year={2024},
  publisher={West Virginia University}
}

@article{cui2023federated,
  title   = {Federated learning with SARIMA-based clustering for carbon emission prediction},
  author  = {Cui, Tianxu and Shi, Ying and Lv, Bo and Ding, Rijia and Li, Xianqiang},
  journal = {Journal of Cleaner Production},
  volume  = {426},
  pages   = {139069},
  year    = {2023},
  publisher = {Elsevier}
}

@article{alkheder2022forecasting,
  title   = {Forecasting of carbon dioxide emissions from power plants in water using United States Environmental Protection Agency, intergovernmental panel on climate change, and machine learning methods},
  author  = {Alkheder, S. and Almusalam, A.},
  journal = {Renewable Energy},
  volume  = {191},
  pages   = {819--827},
  year    = {2022},
  doi     = {10.1016/j.renene.2022.04.023}
}

@inproceedings{alvodji2019iotfla,
  title     = {IOTFLA: a secured and privacy-preserving smart home architecture implementing federated learning},
  author    = {Alvodji, U.M. and Gambs, S. and Martin, A.},
  booktitle = {2019 IEEE Security and Privacy Workshops (SPW)},
  pages     = {175--180},
  year      = {2019},
  doi       = {10.1109/SPW.2019.00041}
}

@article{appiah2018investigating,
  title   = {Investigating the multivariate Granger causality between energy consumption, economic growth and CO2 emissions in Ghana},
  author  = {Appiah, M.O.},
  journal = {Energy Policy},
  volume  = {112},
  pages   = {198--208},
  year    = {2018},
  doi     = {10.1016/j.enpol.2017.10.017}
}

@article{chen2020fedhealth,
  title   = {Fedhealth: a federated transfer learning framework for wearable healthcare},
  author  = {Chen, Y. and Qiu, X. and Wang, J. and others},
  journal = {IEEE Intelligent Systems},
  volume  = {35},
  number  = {4},
  pages   = {83--93},
  year    = {2020},
  doi     = {10.1109/MIS.2020.2988604}
}

@article{dai2018forecasting,
  title   = {Forecasting of energy-related CO2 emissions in China based on GM(1,1) and least squares support vector machine optimized by modified shuffled frog leaping algorithm for sustainability},
  author  = {Dai, S. and Niu, D. and Han, Y.},
  journal = {Sustainability},
  volume  = {10},
  number  = {4},
  pages   = {958},
  year    = {2018},
  doi     = {10.3390/su10040958}
}

@inproceedings{gao2021prediction,
  title     = {Prediction method of green transportation carbon emission in smart city based on gray joint algorithm},
  author    = {Gao, B. and Li, X. and Yu, H.},
  booktitle = {2021 6th International Conference on Smart Grid and Electrical Automation (ICSGEA)},
  pages     = {30--34},
  year      = {2021},
  doi       = {10.1109/ICSGEA53208.2021.00015}
}

@article{gholizadeh2022federated,
  title   = {Federated learning with hyperparameter-based clustering for electrical load forecasting},
  author  = {Gholizadeh, N. and Musilek, P.},
  journal = {Internet of Things},
  volume  = {17},
  pages   = {100470},
  year    = {2022},
  doi     = {10.1016/j.iot.2021.100470}
}

@article{hakak2020framework,
  title   = {A framework for edge-assisted healthcare data analytics using federated learning},
  author  = {Hakak, S. and Ray, S. and Khan, W.Z. and others},
  journal = {2020 IEEE International Conference on Big Data (Big Data)},
  pages   = {3423--3427},
  year    = {2020},
  doi     = {10.1109/BigData50022.2020.9378078}
}

@article{ipcc2018global,
  title   = {Global warming of 1.5°C an IPCC special report on the impacts of global warming of 1.5°C above pre-industrial levels and related global greenhouse gas emission pathways},
  author  = {Intergovernmental Panel on Climate Change (IPCC)},
  journal = {The Context of Strengthening the Global Response to the Threat of Climate Change, Sustainable Development, and Efforts to Eradicate Poverty},
  year    = {2018}
}

@article{kairouz2021advances,
  title   = {Advances and open problems in federated learning},
  author  = {Kairouz, P. and McMahan, H.B. and Avent, B. and others},
  journal = {Foundations and Trends in Machine Learning},
  volume  = {14},
  number  = {1-2},
  pages   = {1--210},
  year    = {2021},
  doi     = {10.1561/2200000083}
}

@article{liu2020privacy,
  title   = {Privacy-preserving traffic flow prediction: a federated learning approach},
  author  = {Liu, Y. and James, J.Q. and Kang, J. and others},
  journal = {IEEE Internet of Things Journal},
  volume  = {7},
  number  = {9},
  pages   = {7751--7763},
  year    = {2020},
  doi     = {10.1109/JIOT.2020.2991401}
}

@article{liu2020carbon,
  title   = {Carbon Monitor, a near-real-time daily dataset of global CO2 emission from fossil fuel and cement production},
  author  = {Liu, Z. and Ciais, P. and Deng, Z. and others},
  journal = {Scientific Data},
  volume  = {7},
  number  = {1},
  pages   = {392},
  year    = {2020},
  doi     = {10.1038/s41597-020-00708-7}
}

@article{liu2024online,
  title={Online spatio-temporal correlation-based federated learning for traffic flow forecasting},
  author={Liu, Qingxiang and Sun, Sheng and Liu, Min and Wang, Yuwei and Gao, Bo},
  journal={IEEE Transactions on Intelligent Transportation Systems},
  year={2024},
  publisher={IEEE}
}

@article{niu2020can,
  title   = {Can China achieve its 2030 carbon emissions commitment? Scenario analysis based on an improved general regression neural network},
  author  = {Niu, D. and Wang, K. and Wu, J. and others},
  journal = {Journal of Cleaner Production},
  volume  = {243},
  pages   = {118558},
  year    = {2020},
  doi     = {10.1016/j.jclepro.2019.118558}
}

@article{ardakani2023federated,
  title   = {A federated learning-enabled predictive analysis to forecast stock market trends},
  author  = {Pourroostaei Ardakani, S. and Du, N. and Lin, C. and others},
  journal = {Journal of Ambient Intelligence and Humanized Computing},
  pages   = {1--7},
  year    = {2023},
  doi     = {10.1007/s12652-023-04570-4}
}

@inproceedings{saputra2019energy,
  title     = {Energy demand prediction with federated learning for electric vehicle networks},
  author    = {Saputra, Y.M. and Hoang, D.T. and Nguyen, D.N. and others},
  booktitle = {2019 IEEE Global Communications Conference (GLOBECOM)},
  pages     = {1--6},
  year      = {2019},
  doi       = {10.1109/GLOBECOM38437.2019.9013587}
}

@inproceedings{siami2019performance,
  title     = {The performance of LSTM and BiLSTM in forecasting time series},
  author    = {Siami-Namini, S. and Tavakoli, N. and Namin, A.S.},
  booktitle = {2019 IEEE International Conference on Big Data (Big Data)},
  pages     = {3285--3292},
  year      = {2019},
  doi       = {10.1109/BigData47090.2019.9005595}
}

@article{sun2022predictions,
  title   = {Predictions of carbon emission intensity based on factor analysis and an improved extreme learning machine from the perspective of carbon emission efficiency},
  author  = {Sun, W. and Huang, C.},
  journal = {Journal of Cleaner Production},
  volume  = {338},
  pages   = {130414},
  year    = {2022},
  doi     = {10.1016/j.jclepro.2022.130414}
}

@inproceedings{tun2021federated,
  title     = {Federated learning based energy demand prediction with clustered aggregation},
  author    = {Tun, Y.L. and Thar, K. and Tiwari, C.M. and others},
  booktitle = {2021 IEEE International Conference on Big Data and Smart Computing (BigComp)},
  pages     = {164--167},
  year      = {2021},
  doi       = {10.1109/BigComp51126.2021.00041}
}

@article{wang2022grey,
  title={Forecasting Chinese provincial carbon emissions using a novel grey prediction model considering spatial correlation},
  author={Wang, Huiping and Zhang, Zhun},
  journal={Expert Systems with Applications},
  volume={209},
  pages={118261},
  year={2022},
  publisher={Elsevier}
}

@article{yu2021fedkd,
  title   = {Fedkd: Communication efficient federated learning via knowledge distillation},
  author  = {Yu, T. and Li, T. and Sun, S. and others},
  journal = {Advances in Neural Information Processing Systems},
  volume  = {34},
  year    = {2021}
}

@article{konevcny2016federated,
  title={Federated learning: Strategies for improving communication efficiency},
  author={Kone{\v{c}}n{\`y}, Jakub and McMahan, H Brendan and Yu, Felix X and Richt{\'a}rik, Peter and Suresh, Ananda Theertha and Bacon, Dave},
  journal={arXiv preprint arXiv:1610.05492},
  year={2016}
}

@inproceedings{zhang2022federated,
  title={Federated learning-based short-term building energy consumption prediction method for solving the data silos problem},
  author={Li, Junyang and Zhang, Chaobo and Zhao, Yang and Qiu, Weikang and Chen, Qi and Zhang, Xuejun},
  booktitle={Building Simulation},
  volume={15},
  number={6},
  pages={1145--1159},
  year={2022},
  organization={Springer}
}

@article{zhao2022china,
  title={Challenges toward carbon neutrality in China: Strategies and countermeasures},
  author={Zhao, Xin and Ma, Xiaowei and Chen, Boyang and Shang, Yuping and Song, Malin},
  journal={Resources, Conservation and Recycling},
  volume={176},
  pages={105959},
  year={2022},
  publisher={Elsevier}
}

@inproceedings{nilsson2018performance,
  title={A performance evaluation of federated learning algorithms},
  author={Nilsson, Adrian and Smith, Simon and Ulm, Gregor and Gustavsson, Emil and Jirstrand, Mats},
  booktitle={Proceedings of the second workshop on distributed infrastructures for deep learning},
  pages={1--8},
  year={2018}
}

@article{sun2022decentralized,
  title={Decentralized federated averaging},
  author={Sun, Tao and Li, Dongsheng and Wang, Bao},
  journal={IEEE Transactions on Pattern Analysis and Machine Intelligence},
  volume={45},
  number={4},
  pages={4289--4301},
  year={2022},
  publisher={IEEE}
}

@ARTICLE{9599369,
  author={Li, Qinbin and Wen, Zeyi and Wu, Zhaomin and Hu, Sixu and Wang, Naibo and Li, Yuan and Liu, Xu and He, Bingsheng},
  journal={IEEE Transactions on Knowledge and Data Engineering}, 
  title={A Survey on Federated Learning Systems: Vision, Hype and Reality for Data Privacy and Protection}, 
  year={2023},
  volume={35},
  number={4},
  pages={3347-3366},
  doi={10.1109/TKDE.2021.3124599}
}

@ARTICLE{10423871,
  author={Yang, Xin and Yu, Hao and Gao, Xin and Wang, Hao and Zhang, Junbo and Li, Tianrui},
  journal={IEEE Transactions on Knowledge and Data Engineering}, 
  title={Federated Continual Learning via Knowledge Fusion: A Survey}, 
  year={2024},
  volume={36},
  number={8},
  pages={3832-3850},
  doi={10.1109/TKDE.2024.3363240}
}

@ARTICLE{9996132,
  author={Zhang, Lefeng and Zhu, Tianqing and Xiong, Ping and Zhou, Wanlei and Yu, Philip S.},
  journal={IEEE Transactions on Knowledge and Data Engineering}, 
  title={A Game-Theoretic Federated Learning Framework for Data Quality Improvement}, 
  year={2023},
  volume={35},
  number={11},
  pages={10952-10966},
  doi={10.1109/TKDE.2022.3230959}
}

\end{document}